\documentclass[letterpaper]{article} 
\usepackage{aaai2026}  
\usepackage{times}  
\usepackage{helvet}  
\usepackage{courier}  
\usepackage[hyphens]{url}  
\usepackage{graphicx} 
\usepackage{natbib}  
\usepackage{caption} 
\usepackage{algorithm}
\usepackage{algorithmic}
\usepackage{amsmath}
\usepackage{amsmath,amssymb}
\usepackage{booktabs} 
\usepackage{multirow}
\usepackage{newfloat}
\usepackage{listings}
\usepackage{hyperref}
\DeclareCaptionStyle{ruled}{labelfont=normalfont,labelsep=colon,strut=off} 
\floatstyle{ruled}
\newfloat{listing}{tb}{lst}{}
\floatname{listing}{Listing}
\usepackage{xcolor}

\title{Inference-Time Alignment Control for Diffusion Models with \\ Reinforcement Learning Guidance}

\author {
    Luozhijie Jin\textsuperscript{\rm 1}\equalcontrib,
    Zijie Qiu\textsuperscript{\rm 1}\equalcontrib,
    Jie Liu\textsuperscript{\rm 3},
    Zijie Diao\textsuperscript{\rm 1},
    Lifeng Qiao\textsuperscript{\rm 4},
    \\
    Ning Ding\textsuperscript{\rm 2}\correspondingauthor,
    Alex Lamb\textsuperscript{\rm 2}\correspondingauthor,
    Xipeng Qiu\textsuperscript{\rm 1}\correspondingauthor
}
\affiliations {
    \textsuperscript{\rm 1}Fudan University\\
    \textsuperscript{\rm 2}Tsinghua University\\
    \textsuperscript{\rm 3}CUHK MMLab\\
    \textsuperscript{\rm 4}Shanghai Jiaotong University\\
    \{lzjjin22, zjqiu22\}@m.fudan.edu.cn\\
    xpqiu@fudan.edu.cn
    lambalex@tsinghua.edu.cn dingning@mail.tsinghua.edu.cn
}

\usepackage{bibentry}
\nocopyright
\begin{document}

\maketitle

\begin{abstract}
Denoising-based generative models, particularly diffusion and flow matching algorithms, have achieved remarkable success. However, aligning their output distributions with complex downstream objectives—such as human preferences, compositional accuracy, or data compressibility—remains challenging. While reinforcement learning (RL) fine-tuning methods, inspired by advances in RL from human feedback (RLHF) for large language models, have been adapted to these generative frameworks, current RL approaches are suboptimal for diffusion models and offer limited flexibility in controlling alignment strength after fine-tuning.
In this work, we reinterpret RL fine-tuning for diffusion models through the lens of stochastic differential equations and implicit reward conditioning. 
We introduce \emph{Reinforcement Learning Guidance} (RLG), an inference-time method that adapts Classifier-Free Guidance (CFG) by combining the outputs of the base and RL fine-tuned models via a geometric average. 
Our theoretical analysis shows that RLG's guidance scale is mathematically equivalent to adjusting the KL-regularization coefficient in standard RL objectives, enabling dynamic control over the alignment-quality trade-off without further training. 
Extensive experiments demonstrate that RLG consistently improves the performance of RL fine-tuned models across various architectures, RL algorithms, and downstream tasks, including human preferences, compositional control, compressibility, and text rendering. Furthermore, RLG supports both interpolation and extrapolation, thereby offering unprecedented flexibility in controlling generative alignment. Our approach provides a practical and theoretically sound solution for enhancing and controlling diffusion model alignment at inference. The source code for RLG is publicly available at the Github:
\url{https://github.com/jinluo12345/Reinforcement-learning-guidance}.

\end{abstract}

\begin{figure*}[h!]
    \centering
    \includegraphics[width=0.8\textwidth]{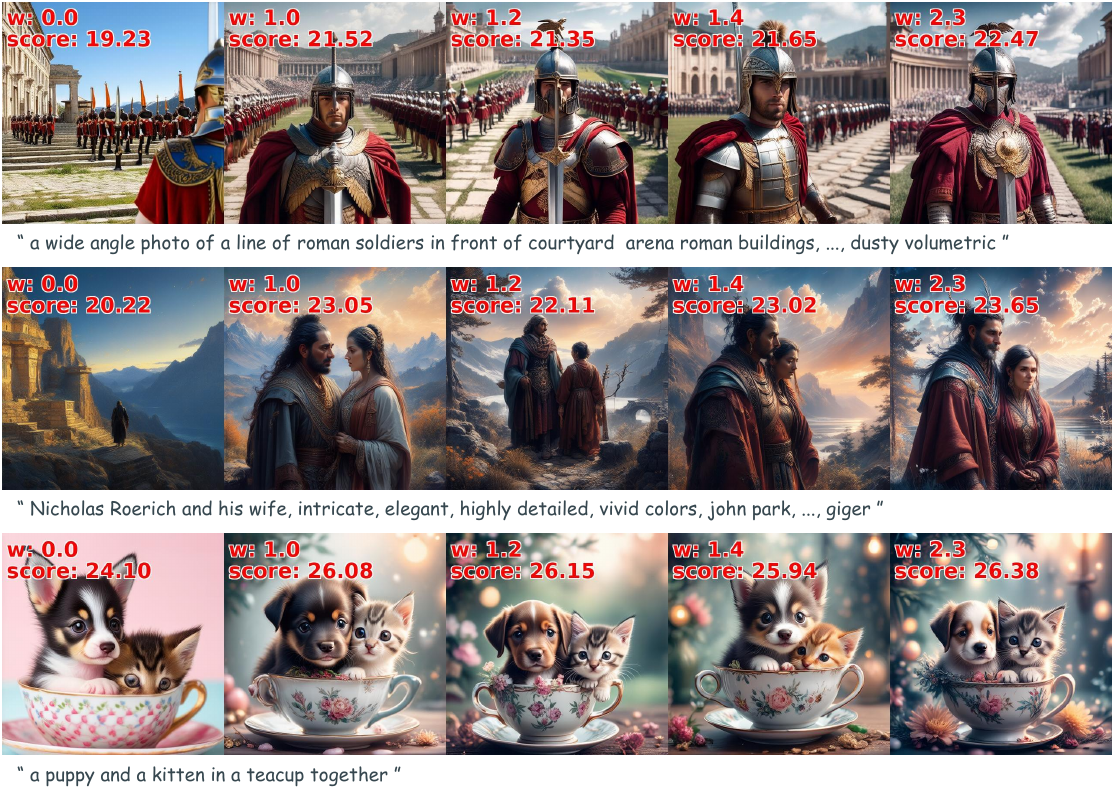}
    \caption{Selected qualitative results for the human preference alignment task using SD3.5-M with GRPO and our RLG. The PickScore is displayed on each image. As the RLG scale increases, the images generally become more detailed, aesthetically pleasing, which is corroborated by the rising PickScores.}
    \label{fig:aes}
\end{figure*}


\section{Introduction}
While denoising-based generative models—primarily diffusion~\cite{ho2020denoising,rombach2022high} and flow matching~\cite{lipman2022flow,esser2024scalingrectifiedflowtransformers} algorithms—have gained widespread acceptance and usage, a new challenge has emerged: aligning the learned distribution with complex downstream objectives. Such alignment may involve making the generative distribution consistent with human preferences~\cite{kirstain2023pickapicopendatasetuser} or ensuring compliance with stringent generation requirements, such as compositional correctness~\cite{ghosh2023geneval}, fidelity in text rendering~\cite{liu2025flow}, or data compressibility~\cite{black2023training}. Existing methods~\citep{liu2024alignment} for aligning denoising-based generative models include reward-weighted regression~\cite{peng1910simple,lee2023aligning,fan2025onlinerewardweightedfinetuningflow}, direct reward fine-tuning~\cite{xu2023imagereward, prabhudesai2023aligning, clark2023directly}, and reinforcement learning (RL) fine-tuning.

Owing to significant advancements in Reinforcement Learning from Human Feedback (RLHF)~\cite{black2023training,lee2023aligning} for Large Language Models (LLMs), an increasing number of researchers are adapting RL fine-tuning techniques from LLMs to denoising-based generative models. This adaptation is primarily achieved by interpreting the iterative denoising process as a multi-step decision-making problem, which facilitates the application of RL algorithms—such as REINFORCE~\cite{williams1992simple, mohamed2020monte, black2023training}, Direct Preference Optimization (DPO)~\cite{rafailov2023direct, wallace2024diffusion}, and Group Relative Policy Optimization (GRPO)~\cite{shao2024deepseekmath, liu2025flow}—to diffusion models. However, current RL methods for diffusion models still exhibit several limitations, primarily in two respects. First, the exact probability of a sampled image is intractable due to the nature of diffusion algorithms, which undermines the effectiveness of existing RL algorithms~\cite{black2023training, gong2025diffucoder}. Second, the degree to which the base model aligns with downstream objectives remains fixed after RL fine-tuning and is sensitive to hyperparameter choices, such as the Kullback–Leibler (KL) coefficient. This inflexibility prevents users from dynamically balancing alignment and generation quality, which may be crucial in applications such as compressibility.

In this work, we draw inspiration from the stochastic differential equation (SDE) nature of denoising-based generative models~\cite{song2020score}, which motivates us to interpret RL fine-tuning of diffusion models as a special case of generation conditioned on implicit rewards learned through reinforcement learning objectives~\cite{rafailov2024rqlanguagemodel, zhu2025dspo, cui2025process}. Building upon this perspective, we introduce an inference-time enhancement technique, \emph{Reinforcement Learning Guidance} (RLG), which adapts the established controlling approach, Classifier-Free Guidance (CFG)~\cite{ho2022classifier, zheng2023guided}, by computing a weighted sum of the outputs from the base model and the RL fine-tuned model using a geometric average. We theoretically demonstrate that this weighted averaging has the same effect as modifying the KL coefficient in RL fine-tuning, but crucially, it requires no additional training.

Empirical results on downstream tasks demonstrate that RLG enhances the performance of RL fine-tuned models across various model types (diffusion and flow matching), a range of RL methods (policy gradient, DPO, GRPO, etc.), and multiple downstream objectives (image aesthetics, compositional control, compressibility, text rendering, inpainting, and personalized generation). Furthermore, RLG supports both interpolation and extrapolation, thereby offering substantial flexibility in controlling the degree of alignment with downstream objectives. 

Our contributions are summarized as follows:
\begin{itemize}
    \item We propose \emph{Reinforcement Learning Guidance} (RLG), a novel, training-free approach for enhancing and controlling the alignment of diffusion models at inference time.
    \item We provide a theoretical foundation for RLG, demonstrating that its guidance scale is mathematically equivalent to adjusting the KL-regularization coefficient in the underlying RL objective. This analysis formally accounts for the effectiveness of extrapolation ($w>1$).
    \item We perform extensive experiments on a diverse set of alignment tasks, showing that RLG consistently enhances performance by enabling models to surpass their original fine-tuned capabilities, while also allowing for flexible trade-offs between competing objectives.
\end{itemize}

\section{Background}

\subsection{Diffusion and Flow-based Generative Models}

Generative modeling has advanced rapidly, particularly through diffusion~\cite{ho2020denoising,song2020denoising,rombach2022high} and flow-based~\cite{lipman2022flow,liu2022flow,esser2024scalingrectifiedflowtransformers} methods. Modern diffusion models, such as DDPM~\cite{ho2020denoising}, DDIM~\cite{song2020denoising}, and Stable Diffusion~\cite{rombach2022high}, generate samples by gradually corrupting data with noise via a forward stochastic process (SDE) and training a neural network to reverse this process, learning the score function to denoise data~\cite{song2020score}. 

Flow-based generative models, such as Flow Matching~\cite{lipman2022flow}, instead directly learn a deterministic trajectory between a simple prior and the data distribution, described by an ordinary differential equation (ODE). This approach, and its recent extensions~\cite{liu2022flow,esser2024scalingrectifiedflowtransformers,tong2023improving,kong2024hunyuanvideo,liu2025lumina,wan2025wan}, has led to faster convergence and more stable training.

Formally, consider a reference flow $(X_t)_{t \in [0,1]}$ interpolating between noise $X_1 \sim p_1$ and data $X_0 \sim p_{\text{data}}$, commonly parameterized as:
\begin{equation} \label{eq:prob_path}
    X_t = \beta_t X_1 + \alpha_t X_0 ,
\end{equation}
with schedules $\alpha_0 = \beta_1 = 0$, $\alpha_1 = \beta_0 = 1$. A typical choice is $\alpha_t = t$, $\beta_t = 1-t$.

\textbf{Flow Matching} models learn a velocity field $\mathbf{v}(\mathbf{X}_t, t)$, resulting in an ODE: $d\mathbf{X}_t = \mathbf{v}(\mathbf{X}_t, t) dt$. The model is trained to match the velocity of the reference flow:
\[
    \mathbf{v}(\mathbf{X}_t, t) \approx \frac{d}{dt} X_t = \dot{\beta}_t X_0 + \dot{\alpha}_t X_1.
\]

\textbf{Diffusion Models} rely on a stochastic differential equation (SDE), with the model predicting the score function $\mathbf{s}(\mathbf{X}_t, t) = \nabla_{\mathbf{x}} \log p_t(\mathbf{X}_t)$. The interpolation schedules are typically $\alpha_t = \sqrt{\bar{\alpha}_t}$ and $\beta_t = \sqrt{1 - \bar{\alpha}_t}$, where $\bar{\alpha}_t$ increases from $0$ to $1$.

The connection between the two approaches can be formalized: the velocity field in Flow Matching can be written in terms of the score function as
\begin{equation} \label{eq:v2s}
    \mathbf{v}(\mathbf{x}, t) = \left( \frac{\dot{\alpha}_t}{\alpha_t} \right) \mathbf{x} + \beta_t \left( \frac{\dot{\alpha}_t}{\alpha_t} \beta_t - \dot{\beta}_t \right) \mathbf{s}(\mathbf{x}, t).
\end{equation}

Both paradigms can be unified in the framework of a general SDE:
\begin{equation} \label{eq:reverse_sde}
    d\mathbf{X}_t = \left(\mathbf{v}(\mathbf{X}_t,t) - \frac{1}{2}\sigma(t)^2\mathbf{s}(\mathbf{X}_t,t)\right)dt + \sigma(t) dw
\end{equation}
where $w$ is standard Brownian motion, and $\sigma(t)$ determines the stochasticity. Flow Matching and diffusion models are thus specific instances, distinguished by their velocity field, score function, and noise schedule.

\subsection{Guidance and Control in Generative Models}

Controlling generative model outputs is essential for conditional generation tasks. Early work such as Classifier Guidance (CG) steers the generation process using gradients from a separately trained classifier~\cite{dhariwal2021diffusion}, but this approach is computationally costly and limited by the need for an external model.

Classifier-Free Guidance (CFG) has become the standard alternative~\cite{ho2022classifier}. At inference, CFG computes two passes: one with the actual condition $c$ (e.g., text-guided) and one with the null condition $\emptyset$. The guided velocity field $\hat{\mathbf{v}}_{\theta}$ is a linear interpolation between these two outputs:
\begin{equation} \label{eq:vcfg}
    \hat{\mathbf{v}}_{\theta}(\mathbf{x}_t, t | c) \triangleq (1 - \omega) \mathbf{v}_{\theta}(\mathbf{x}_t, t | \emptyset) + \omega \mathbf{v}_{\theta}(\mathbf{x}_t, t | c),
\end{equation}
where $\omega$ is the guidance scale parameter. Setting $\omega = 1$ recovers conditional generation, while $\omega>1$ extrapolates beyond the conditional prediction.

The same principle applies to the model's score function:
\begin{equation} \label{eq:scfg}
    \hat{\mathbf{s}}_{\theta}(\mathbf{x}_t, t | c) \triangleq (1 - \omega) \mathbf{s}_{\theta}(\mathbf{x}_t, t | \emptyset) + \omega \mathbf{s}_{\theta}(\mathbf{x}_t, t | c),
\end{equation}
where $\mathbf{s}_{\theta}(\mathbf{x}_t, t | c) = \nabla_{\mathbf{x}_t} \log p_{\theta}(\mathbf{x}_t | c)$. This can be equivalently written as:
\begin{equation} \label{eq:cfgscore}
    \hat{\mathbf{s}}_{\theta}(\mathbf{x}_t, t | c) = \nabla_{\mathbf{x}_t} \log \left( p_{\theta}(\mathbf{x}_t)^{1-\omega} p_{\theta}(\mathbf{x}_t | c)^{\omega} \right),
\end{equation}

Although CFG is highly effective, its performance depends sensitively on the static guidance scale $\omega$~\cite{bradley2024classifier}. Recent advances have explored dynamic or adaptive guidance~\cite{sadat2024eliminating}, as well as improved sampling and guidance strategies~\cite{chung2024cfg++,fan2025cfg,karras2024guiding} to address these shortcomings. However, most existing methods focus on adherence to training-time conditions, leaving open the possibility of leveraging reinforcement learning rewards as dynamic, flexible forms of guidance.

\subsection{Preference Alignment in Generative Models}

Multiple recent methods directly adopt preference learning methods originally designed for LLMs to fine-tune T2I diffusion models to align with human preferences.
In this setting, a pre-trained generative model (policy) $\pi_{\text{ref}}$ is fine-tuned to maximize a reward function $R(\mathbf{x})$. The RL objective typically balances reward maximization with a regularization term:
\begin{equation} \label{eq:rl_objective}
    \pi_{\theta}^* = \arg\max_{\pi_{\theta}} \mathbb{E}_{\mathbf{x} \sim \pi_{\theta}(\mathbf{x})} [R(\mathbf{x})] - \beta D_{\text{KL}}(\pi_{\theta}(\mathbf{x}) \| \pi_{\text{ref}}(\mathbf{x})),
\end{equation}
where $\beta$ controls the trade-off between reward and KL regularization.

The optimal solution to this problem has a closed-form expression: the aligned policy is a re-weighted version of the reference policy, where the weights are exponentially proportional to the reward~\cite{peng1910simple,lee2023aligning,fan2025onlinerewardweightedfinetuningflow}:
\begin{equation} \label{eq:rl_dist}
    p^*(\mathbf{x}) \propto p_{\text{ref}}(\mathbf{x}) \exp\left(\frac{1}{\beta} R(\mathbf{x})\right).
\end{equation}
While this objective can be optimized with policy gradient methods like PPO~\cite{schulman2017proximal, black2023training}, recent work has shifted towards more direct approaches like Direct Preference Optimization (DPO)~\cite{rafailov2023direct, wallace2024diffusion} Building on this, Group Relative Policy Optimization (GRPO)~\cite{sun2025f5r, shao2024deepseekmath} further simplifies the process by estimating sample advantages implicitly within a batch.

However, directly adapting these algorithms from LLM domains results in an estimated loss. For instance, Diffusion-DPO~\cite{wallace2024diffusion} is upper-bounded by the original DPO loss. This estimation can lead to suboptimal performance when fine-tuning diffusion models to align with human preferences~\cite{zhu2025dspo}. To tackle this, \cite{zhu2025dspo} draw preference rewards into score matching objectives. However, this can only act on pair data and offers no flexible control.

\paragraph{Concurrent Work.} While finalizing our paper, two concurrent works, CFGRL\citep{frans2025diffusion} and Diffusion Blend\citep{cheng2025diffusion}, appeared. Both investigate inference-time manipulation techniques via score interpolation. However, CFGRL focuses solely on offline RL and simple task settings, while Diffusion Blend does not establish a connection between interpolation and implicit reward guidance. In contrast, RLG offers a comprehensive analysis of score interpolation from the perspective of implicit classifier guidance and demonstrates its effectiveness across various image generation models, RL algorithms, and tasks.

\section{Methods}
We propose Reinforcement Learning Guidance (RLG), a method that re-purposes a model fine-tuned with reinforcement learning to act as a dynamic guide during the generative process. 

\subsection*{Deriving Reinforcement Learning Guidance (RLG)}
Let $r$ represent the desired attribute, such as a high preference score. Following Bayes' rule, the score function of the conditional distribution $p_{\text{ref}}(\mathbf{x}_t | r)$ can be decomposed as:
\begin{equation}
    \nabla_{\mathbf{x}_t} \log p_{\text{ref}}(\mathbf{x}_t | r) = \nabla_{\mathbf{x}_t} \log p_{\text{ref}}(\mathbf{x}_t) + \nabla_{\mathbf{x}_t} \log p(r | \mathbf{x}_t).
    \label{eq:score_decomposition}
\end{equation}

To relate this to a reward function $R(\mathbf{x}_t)$, we can model the likelihood $p(r | \mathbf{x}_t)$ using an energy-based formulation according to~\cite{zhu2025dspo}. In this formulation, states with higher rewards are exponentially more likely. Specifically, $p(r | \mathbf{x}_t)$ is defined as:
\begin{equation}
    p(r | \mathbf{x}_t) = \frac{\exp(R(\mathbf{x}_t))}{Z},
\end{equation}
where $Z = \int \exp(R(\mathbf{x}_t)) d\mathbf{x}_t$ is the normalization constant that ensures the probabilities sum or integrate to one over the entire data space.

Substituting this into Equation~\ref{eq:score_decomposition}, the log cancels the exponential, and we obtain the general formula for reward gradient guidance:
\begin{equation}
    \hat{\mathbf{s}}(\mathbf{x}_t, t) = \mathbf{s}_{\text{ref}}(\mathbf{x}_t, t) + \eta \nabla_{\mathbf{x}_t} R(\mathbf{x}_t).
\end{equation}

Here, $\eta$ is a guidance scale. In our setting, we do not have an explicit, differentiable reward model $R(\mathbf{x}_t)$. Drawing from the solution to the KL-regularized RL objective (Equation \ref{eq:rl_dist}) and from~\cite{rafailov2024rqlanguagemodel,zhu2025dspo}, we can reverse-engineer and define an implicit, time-dependent reward function $R_t(\mathbf{x}_t)$ that represents the preference learned by $\pi_\theta$ at each point in the generative process:
\begin{equation}
    R_t(\mathbf{x}_t) \triangleq \beta \log \frac{p_{\theta,t}(\mathbf{x}_t)}{p_{\text{ref},t}(\mathbf{x}_t)}.
    \label{eq:implicit_reward_def}
\end{equation}

Here, $p_{\theta,t}$ and $p_{\text{ref},t}$ are the marginal probability distributions of the noisy sample $\mathbf{x}_t$ under the RL-aligned and reference models, respectively, and $\beta$ is the KL-coefficient from the original RL fine-tuning objective. 

To use our implicit reward for guidance, we need its gradient. Taking the gradient of Equation \ref{eq:implicit_reward_def} with respect to $\mathbf{x}_t$ yields a remarkably simple result:
\begin{align}
    \nabla_{\mathbf{x}_t} R_t(\mathbf{x}_t) &= \nabla_{\mathbf{x}_t} \left( \beta \log \frac{p_{\theta,t}(\mathbf{x}_t)}{p_{\text{ref},t}(\mathbf{x}_t)} \right) \nonumber \\
    &= \beta \left( \nabla_{\mathbf{x}_t} \log p_{\theta,t}(\mathbf{x}_t) - \nabla_{\mathbf{x}_t} \log p_{\text{ref},t}(\mathbf{x}_t) \right) \nonumber \\
    &= \beta \left( \mathbf{s}_{\theta}(\mathbf{x}_t, t) - \mathbf{s}_{\text{ref}}(\mathbf{x}_t, t) \right).
\end{align}

Substituting this gradient into the general reward guidance formula and defining a single, user-controlled guidance scale $w \triangleq \eta\beta$, we arrive at the score-based formulation of RLG:
\begin{align}
    \hat{\mathbf{s}}_{\text{RLG}}(\mathbf{x}_t, t) &= \mathbf{s}_{\text{ref}} + w (\mathbf{s}_{\theta} - \mathbf{s}_{\text{ref}}) \nonumber \\
    &= (1-w)\mathbf{s}_{\text{ref}} + w \mathbf{s}_{\theta}.
\end{align}

This demonstrates that the linear interpolation of score functions is a direct implementation of implicit reward gradient guidance. Using the velocity-score relationship from Equation \ref{eq:v2s}, the same linear interpolation applies to the velocity fields. This leads to our proposed method, \textbf{Reinforcement Learning Guidance (RLG)}, defined by the guided velocity field:
\begin{equation} \label{eq:rlg_velocity}
    \hat{\mathbf{v}}_{\text{RLG}}(\mathbf{x}_t, t) \triangleq (1-w)\mathbf{v}_{\text{ref}}(\mathbf{x}_t, t) + w \mathbf{v}_{\theta}(\mathbf{x}_t, t),
\end{equation}
where $w$ is the RLG guidance scale. A value of $w=0$ recovers the original model, $w=1$ recovers the RL-finetuned model, and $w>1$ extrapolates the learned alignment. The full sampling procedure is outlined in Algorithm \ref{alg:rlg_sampling}.

\begin{algorithm}[tb]
\caption{Sampling with Reinforcement Learning Guidance (RLG)}
\label{alg:rlg_sampling}
\begin{algorithmic}[1]
    \STATE \textbf{Input:} Pre-trained model velocity $\mathbf{v}_{\text{ref}}$, RL-finetuned model velocity $\mathbf{v}_{\theta}$, condition $c$, RLG scale $w$, number of steps $N$.
    \STATE Sample initial noise $\mathbf{x}_1 \sim \mathcal{N}(0, \mathbf{I})$.
    \FOR{$t=1, \dots, N$}
        \STATE Compute reference velocity: $\mathbf{v}_{\text{ref}, t} = \mathbf{v}_{\text{ref}}(\mathbf{x}_t, t | c)$.
        \STATE Compute RL-aligned velocity: $\mathbf{v}_{\text{RL}, t} = \mathbf{v}_{\text{RL}}(\mathbf{x}_t, t | c)$.
        \STATE Compute the guided velocity using RLG:
        \STATE $\hat{\mathbf{v}}_{\text{RLG}, t} = (1-w)\mathbf{v}_{\text{ref}, t} + w\mathbf{v}_{\text{RL}, t}$.
        \STATE Update the sample using a chosen ODE solver step:
        \STATE $\mathbf{x}_{t+1} = \text{SolverStep}(\mathbf{x}_t, \hat{\mathbf{v}}_{\text{RLG}, t})$.
    \ENDFOR
    \STATE \textbf{Return:} Generated sample $\mathbf{x}_{N+1}$.
\end{algorithmic}
\end{algorithm}

\subsection*{Theoretical Justification: RLG as KL-Coefficient Control}

We can now provide a complementary theoretical justification for RLG that explains its mechanisms. Similar to CFG~\cite{ho2022classifier}, the guided score $\hat{\mathbf{s}}_{\text{RLG}}$ corresponds to sampling from a new time-dependent distribution $\hat{p}_{\text{RLG},t}$:
\begin{equation}
    \hat{\mathbf{s}}_{\text{RLG}} = \nabla_{\mathbf{x}_t} \log \left( p_{\text{ref},t}(\mathbf{x}_t)^{1-w} p_{\theta,t}(\mathbf{x}_t)^{w} \right).
\end{equation}

As $t \to 0$. In this limit, the noisy sample $\mathbf{x}_t$ approaches the clean data $\mathbf{x}_0$, and the marginal distributions $p_{\text{ref},t}$ and $p_{\theta,t}$ converge to their corresponding final distributions, $p_{\text{ref}}(\mathbf{x}_0)$ and $p_{\theta}(\mathbf{x}_0)$. Therefore, the score function guiding the final steps of generation points towards a target distribution $\hat{p}_{\text{RLG}}(\mathbf{x}_0)$ of the form:$p_{\text{ref}}(\mathbf{x}_0)^{1-w} p_{\theta}(\mathbf{x}_0)^{w}.$

Assuming the RL-finetuned model $\pi_{\theta}$ has converged to the optimal distribution from~\cite{rafailov2024rqlanguagemodel,rafailov2023direct} (i.e., $p_{\theta}(\mathbf{x}_0) \propto p_{\text{ref}}(\mathbf{x}_0) \exp(\frac{1}{\beta} R(\mathbf{x}_0))$), we can substitute this into the expression for the RLG distribution:
\begin{align}
\label{eq:rlg distribution}
    \hat{p}_{\text{RLG}}(\mathbf{x}_0) &\propto p_{\text{ref}}(\mathbf{x}_0)^{1-w} \left( p_{\text{ref}}(\mathbf{x}_0) \exp\left(\frac{1}{\beta} R(\mathbf{x}_0)\right) \right)^w \nonumber \\
    &\propto p_{\text{ref}}(\mathbf{x}_0) \exp\left(\frac{1}{\beta/w} R(\mathbf{x}_0)\right).
\end{align}

This derivation reveals a crucial insight: applying RLG with a guidance scale $w$ is mathematically equivalent to sampling from the optimal policy of an RL objective with a new, effective KL-regularization coefficient of $\beta/w$. 

\begin{figure*}[h!]
    \centering
    \includegraphics[width=0.85\textwidth]{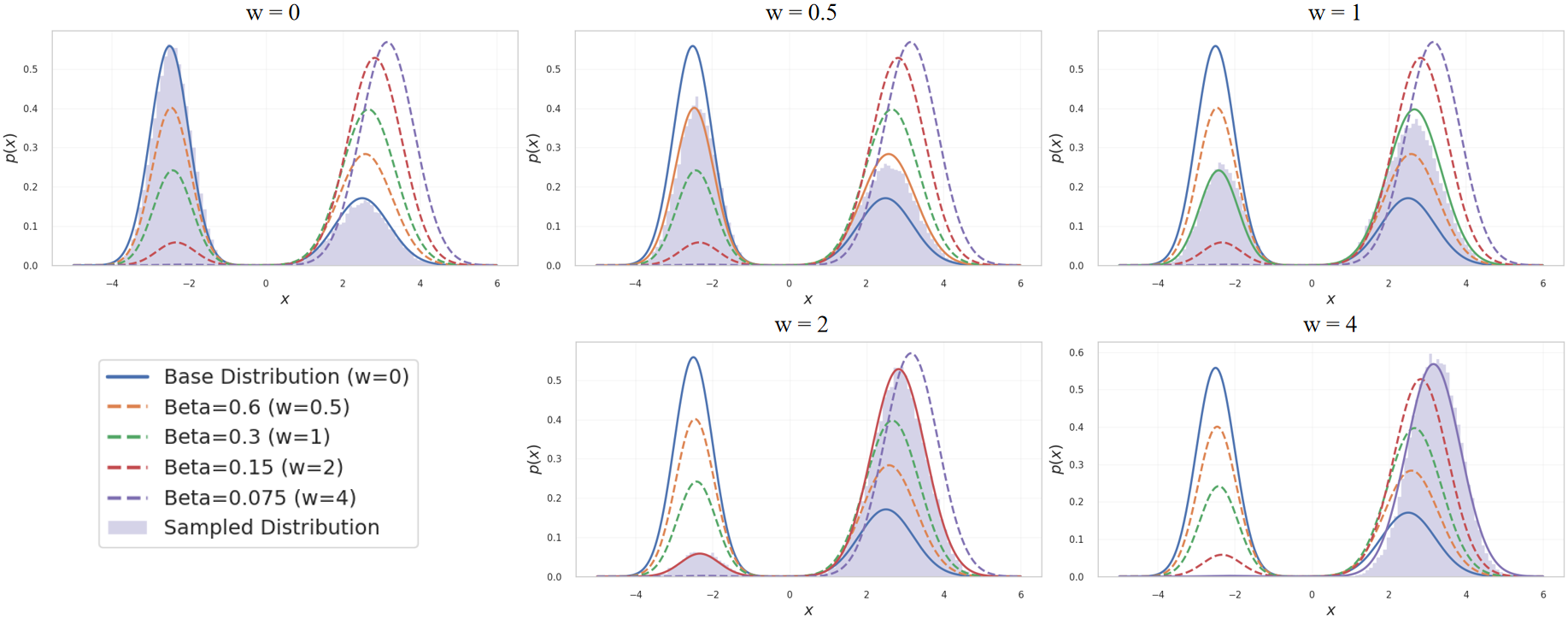}
    \caption{Small-scale demonstration supporting the theoretical justification of RLG. Each subplot shows the sampled distribution under a different RLG weight $w$, while the curves represent the corresponding theoretically predicted RL-fine-tuned distributions. Here, $\beta$ denotes the KL regularization coefficient.}
    \label{fig:tinycase}
\end{figure*}

To empirically validate this result, we conducted a small-scale demonstration. The experimental setting consists of a flow matching model defined on the real line, where the pretrained target (base) distribution is a Gaussian mixture, $p_{\text{base}}(x) \triangleq 0.7 \mathcal{N}(-2.5, 0.25) + 0.3 \mathcal{N}(2.5, 0.49)$. The reward function is defined as $r(x) = 0.1x$. We fine-tuned the pretrained model using the policy gradient algorithm with a KL coefficient $\beta = 0.3$, a batch size of 64, and a learning rate of $1\times 10^{-5}$. Figure~\ref{fig:tinycase} presents the sampled distributions under various RLG weights $w$, alongside the corresponding theoretical distribution curves $p_{\text{rl}}(x) \propto p_{\text{base}}(x) \exp\left(\frac{1}{\beta} r(x)\right)$. The results demonstrate that the distributions sampled with RLG closely match the theoretically predicted RL targets, thereby corroborating our theoretical analysis.

By selecting $w > 1$, we dynamically reduce the regularization penalty at inference time, enabling the model to pursue higher rewards more aggressively than the original RL-fine-tuned model. Conversely, choosing $w < 1$ increases the regularization effect. This provides a principled justification for RLG's capacity to extrapolate or interpolate beyond the learned policy, offering a powerful and theoretically grounded mechanism to control the trade-off between alignment and fidelity.

\section{Experiments}

In this section, we empirically validate the effectiveness of our proposed Reinforcement Learning Guidance (RLG) method. We conduct experiments on various alignment tasks for text-to-image (T2I) models. In each case, the original pre-trained model serves as $\mathbf{v}_{\text{ref}}$ and the RL-aligned model serves as $\mathbf{v}_{\theta}$. We apply RLG as described in Equation~\ref{eq:rlg_velocity}, For all experiments, the number of sampling steps was set to 20. A comprehensive list of additional hyperparameters can be found in the appendix~\ref{sec:appendix_implementation}.
\begin{table*}[t!]
\centering

\begin{tabular}{l|c|cccc}
\toprule
\textbf{Model} & \textbf{RL} & \textbf{$w_{\text{RL}}$} & \textbf{Aesthetic Score ($\uparrow$)} & \textbf{ImageReward ($\uparrow$)} & \textbf{PickScore ($\uparrow$)} \\
\midrule
\multirow{4}{*}{\textbf{SD1.5}} & \multirow{4}{*}{\textbf{DPO}~\cite{lee2023aligning}} & 0.0 & 5.51 / 38.48\% & -0.02 / 36.67\% & 20.03 / 27.34\%\\
& & 1.0 & 5.61 & 0.20 & 20.39 \\
& & 1.4 & 5.63 / 53.56\% & 0.25 / 53.47\%& 20.46 / 57.86\%\\
& & 2.4 & \textbf{5.64} / \textbf{56.25\%}& \textbf{0.32} / \textbf{57.08\%} & \textbf{20.56} / \textbf{61.23\%} \\
\midrule
\multirow{4}{*}{\textbf{SDXL-base}} & \multirow{4}{*}{\textbf{SPO}~\cite{liang2024step}} & 0.0 & 6.10 / 17.87\% & 0.72 / 18.02\% & 21.66 / 7.62\%\\
& & 1.0 & 6.42 & 1.12 & 22.69 \\
& & 1.2 & 6.45 / 59.81\% & 1.13 / 54.10\% &\textbf{22.71} / \textbf{54.64}\%\\
& & 1.4 & \textbf{6.48} / \textbf{62.99}\% & \textbf{1.14} / \textbf{54.15}\% & 22.71 / 54.35\% \\
\midrule
\multirow{4}{*}{\textbf{SD3.5-M}} & \multirow{4}{*}{\textbf{GRPO}~\cite{liu2025flow}} & 0.0 & 5.97 / 11.33\% & 0.99 / 17.29\%& 21.75 / 2.39\% \\
& & 1.0 & 6.45 & 1.40 & 23.29 \\
& & 1.4 & 6.54 / 69.28\% & \textbf{1.40} / \textbf{54.44\%} & 23.48 / \textbf{74.95\%} \\
& & 2.2 & \textbf{6.64} / \textbf{77.39\%} & 1.39 / 53.56\% & \textbf{23.58} / 73.68\% \\
\bottomrule
\end{tabular}
\caption{Quantitative results for Human Preference Alignment. We report mean scores for Aesthetic Score, ImageReward, and PickScore. For each metric, values after the slash (/) denote the win rate (\%) against the standard RL-finetuned model ($w_{\text{RL}}=1.0$). SD denotes stable-diffusion models~\cite{rombach2022high}.}
\label{tab:aesthetic_alignment}
\end{table*}

\subsection{RLG Universally Enhances Alignment Across Diverse Tasks}

A key strength of RLG is its broad applicability. We demonstrate that this training-free method consistently enhances model capabilities across a variety of specialized alignment tasks, ranging from high-level compositional understanding to fine-grained subject fidelity.

\paragraph{Structured Generation: Compositionality and Text Rendering.} We first evaluate RLG on tasks requiring precise adherence to structured prompts. 
For compositional generation, we use the GenEval benchmark~\cite{ghosh2023geneval} with a GRPO-finetuned SD3.5-M model~\cite{liu2025flow}. This task tests the model's ability to correctly render object relationships, counts, and attributes. Following the benchmark's protocol, we use a Mask2Former model~\cite{cheng2022maskedattentionmasktransformeruniversal} to verify the presence and properties of objects specified in the prompts in the official GenEval test set.

For visual text rendering, we use an GRPO-finetuned SD3.5-M model on the Optical Character Recognition (OCR) task~\cite{mori1999optical}, also from \cite{liu2025flow}, to measure its ability to accurately render text from prompts contains the exact string that should appear in the image. We calculate OCR accuracy based on the normalized edit distance, defined as $1 - d_{\text{norm}}$, where $d_{\text{norm}}$ is the Levenshtein distance~\cite{yujian2007normalized} between the generated text extracted from the images using PaddleOCR~\cite{paddleocr2020} and the ground-truth text, normalized by the length of the ground-truth text.

As shown in Table~\ref{tab:geneval_full_results}, \ref{tab:ocr_results} and Figure~\ref{fig:geneval_qualitative}, \ref{fig:ocr_qualitative}, while the RL-finetuned models ($w_{\text{RL}}=1.0$) already show substantial improvements over their base counterparts. By extrapolating with RLG, we unlock further significant gains. On GenEval, RLG pushes the overall score from 93.20\% to a peak of \textbf{94.35\%} and achieves improvement on almost all key compositional tasks. On the OCR task, RLG boosts accuracy from 88.6\% to a new state-of-the-art of \textbf{93.0\%} without much impact on the aesthetic score. These results confirm that RLG can effectively amplify the model's learned ability to follow complex structural constraints.

\begin{table*}[t]
\centering

\begin{tabular}{l|cccccccc}
\toprule
\textbf{Model} &\textbf{$w_{\text{RL}}$} & \textbf{Single Obj.} & \textbf{Two Objs.} & \textbf{Colors} & \textbf{Color Attr.} & \textbf{Counting} & \textbf{Position} & \textbf{Overall Score} \\
\midrule
\multirow{5}{*}{\textbf{SD3.5-M}} & 0.0  & 97.81 & 80.56 & 80.05 & 51.75 & 53.44 & 23.00 & 64.44 \\
&1.0  & \textbf{100.00} & 98.99 & 89.63 & 84.00 & 92.81 & 93.75 & 93.20 \\
&1.2 & 99.69 & 98.74 & 90.69 & 86.11 & 92.81 & 94.25 & 93.72 \\
&1.4 & 99.69 & \textbf{99.24} & 91.49 & \textbf{86.50} & \textbf{94.69} & 94.50 & \textbf{94.35} \\
&1.6 & \textbf{100.00} & 98.99 & \textbf{91.76} & 86.00 & 93.75 & \textbf{95.00} & 94.25 \\
\bottomrule
\end{tabular}
\caption{Pperformance on the GenEval benchmark. We report accuracy (\%) for each compositional sub-task and the overall average score across different RLG guidance scales ($w_{\text{RL}}$).}
\label{tab:geneval_full_results}
\end{table*}

\begin{table}[h!]
\centering

\begin{tabular}{l|ccc}
\toprule
\textbf{Model}& \textbf{$w_{\text{RL}}$} & \textbf{OCR Acc ($\uparrow$)} & \textbf{Aesthetic Score ($\uparrow$)}\\
\midrule
\multirow{6}{*}{SD3.5-M}& 0.0 & 0.543 & 5.40\\
&0.4 & 0.785 & 5.28\\
&0.6 & 0.838 & 5.25\\
&1.0 & 0.886 & 5.20\\
&1.2 & 0.894 & 5.17\\
&1.6 & 0.910 & 5.13\\
&2.2 & 0.921 & 5.07\\
&2.8 & \textbf{0.930} & 5.00\\
\bottomrule
\end{tabular}
\caption{Quantitative results for the visual text rendering task. This table shows the Optical Character Recognition (OCR) accuracy at different RLG guidance scales.}
\label{tab:ocr_results}
\end{table}

\paragraph{Fidelity-Driven Generation: Inpainting and Personalization.} We next test RLG on tasks demanding high fidelity to reference content. We choose image inpainting task and personalized generation task.
For image inpainting, we use PrefPaint~\cite{bui2025prefpaintenhancingimageinpainting} model, an RL-finetuned model designed to fill masked regions according to human preferences, whose base model is stable-diffusion-inpainting~\cite{podell2023sdxlimprovinglatentdiffusion}
To evaluate the quality of the inpainted images, we use Preference Reward~\cite{bui2025prefpaintenhancingimageinpainting} metrics evaludated on dataset provided in the PrefPaint study, which are detailed in appendix~\ref{sec:appendix_inpainting}.
For personalized generation, we use PatchDPO~\cite{huang2025patchdpo}, an RL-finetuned model optimized to maintain the identity of a subject from reference images. We treat the original pre-trained model (IP-Adapter-Plus~\cite{ye2023ip}) as the base model ($\mathbf{v}_\text{ref}$) and PatchDPO as the RL-aligned model ($\mathbf{v}_\theta$). To measure the fidelity of the generated images to the reference subject, we use two standard image-similarity metrics:CLIP-I~\cite{ruiz2023dreambooth} and DINO~\cite{caron2021emerging}, and evaluate the performance on the standard DreamBench~\cite{ruiz2023dreambooth} benchmark, which are detailed in appendix~\ref{sec:appendix_personalize}

The results, summarized in Table~\ref{tab:fidelity_section}, again show RLG's effectiveness. For inpainting, RLG pushes the preference score beyond the original PrefPaint model, peaking at \textbf{0.368}. For personalized generation, RLG further refines subject fidelity, increasing the DINO score to \textbf{0.730} and CLIP-I score to \textbf{0.843}. In both cases, RLG provides a measurable enhancement over state-of-the-art RL-finetuned models without any additional trainings.

\begin{table}[h!]
\centering

\setlength{\tabcolsep}{2.5pt}

\begin{tabular}{l c c} 
\toprule
\multicolumn{3}{l}{\textit{\textbf{Task: Image Inpainting}}} \\

\textbf{Method} & \multicolumn{2}{c}{\textbf{Preference Reward ($\uparrow$)}} \\ 
\cmidrule(r){1-1} \cmidrule(l){2-3} 
\quad Base ($w_\text{RL}=0$)      & \multicolumn{2}{c}{0.080} \\
\quad PrefPaint ($w_\text{RL}=1.0$)  & \multicolumn{2}{c}{0.358} \\ 
\quad RLG ($w_\text{RL}=1.2$)    & \multicolumn{2}{c}{0.367} \\ 
\quad RLG ($w_\text{RL}=1.4$)    & \multicolumn{2}{c}{\textbf{0.368}} \\ 
\quad RLG ($w_\text{RL}=1.6$)    & \multicolumn{2}{c}{0.366} \\ 

\midrule 
\multicolumn{3}{l}{\textit{\textbf{Task: Personalized Generation}}} \\
\textbf{Method} & \textbf{DINO ($\uparrow$)} & \textbf{CLIP-I ($\uparrow$)} \\ 
\cmidrule(r){1-1} \cmidrule(lr){2-2} \cmidrule(l){3-3}
\quad IP-Adapter-Plus ($w_\text{RL}=0$)            & 0.692 & 0.826 \\
\quad PatchDPO ($w_\text{RL}=1.0$)           & 0.724 & 0.839 \\ 
\quad RLG ($w=1.2$) & \textbf{0.730} & 0.841 \\
\quad RLG ($w=1.8$) & 0.730 & \textbf{0.843} \\
\bottomrule
\end{tabular}
\caption{RLG enhances performance on distinct fidelity-driven tasks. The evaluation metrics are presented separately for each task.}
\label{tab:fidelity_section}
\end{table}

\subsection{RLG is Effective Across Diverse RL Algorithms and Model Architectures}

To demonstrate its broad applicability and its model-agnostic nature, we evaluate whether RLG can consistently enhance models that differ in both their underlying generative architecture (i.e., standard diffusion vs. modern flow matching) and the specific reinforcement learning algorithm used for their initial alignment. 
This also extends to algorithms such as GRPO, whose optimal policy does not necessarily conform to Equation~\ref{eq:rl_dist}.

\paragraph{Experimental Setup.}
We conduct this analysis on the human preference alignment task, leveraging three distinct, publicly available RL-finetuned models, each representing a unique combination of architecture and alignment method:
\begin{itemize}
    \item \textbf{SD1.5 + DPO:} A Stable Diffusion v1.5 model~\cite{rombach2022high} aligned using Direct Preference Optimization (DPO)~\cite{wallace2024diffusion}.
    \item \textbf{SDXL + SPO:} A Stable Diffusion XL model~\cite{podell2023sdxlimprovinglatentdiffusion} aligned using Step-wise Preference Optimization (SPO)~\cite{liang2024step}.
    \item \textbf{SD3.5-M + GRPO:} A Stable Diffusion 3.5 Medium flow matching model~\cite{esser2024scalingrectifiedflowtransformers} aligned using Group Relative Policy Optimization (GRPO)~\cite{liu2025flow}.
\end{itemize}
For evaluation, we use three established automated reward models: Aesthetic Score, ImageReward, and PickScore, with details provided in the appendix.

\paragraph{Results.}
The quantitative results, summarized in Table~\ref{tab:aesthetic_alignment}, unequivocally demonstrate that RLG consistently delivers a significant performance boost across all configurations. The effect is particularly pronounced on the state-of-the-art GRPO-tuned SD3.5-M flow model, where RLG achieves a win rate of \textbf{74.95\%} on PickScore against the original finetuned model ($w_{\text{RL}}=1.0$). As visually confirmed in Figures~\ref{fig:aes}, \ref{fig:SD3.5_qualitative}, \ref{fig:SD1.5_qualitative} and \ref{fig:SDXL_qualitative}, increasing the RLG scale consistently enhances image detail and aesthetic appeal.

\subsection{RLG Enables Flexible Control Over Alignment Strength}

A significant limitation of standard RL fine-tuning is that the strength of alignment is permanently baked into the final model. This offers no flexibility for users at inference time. In contrast, RLG provides a powerful, training-free mechanism to dynamically control the degree of alignment.
\paragraph{Controlling a Fundamental Property: Image Compressibility.}

We first demonstrate RLG's control over a low-level property: image compressibility. We used two models fine-tuned with DDPO~\cite{black2023training} with SD1.4 as base model to reward either high or low image compressibility. Standard RL produces a model with a fixed alignment; for instance, the low-compressibility model is locked into a 1.35 compression ratio (where $w_{\text{RL}}=1.0$). RLG transforms this static point into a dynamic spectrum. As shown in Table~\ref{tab:compressibility_tradeoff}, users can weaken the alignment by setting $w_{\text{RL}}<1.0$ (e.g., achieving a 1.14 ratio) or intensify it by setting $w_{\text{RL}}>1.0$, pushing the ratio beyond the fine-tuned limit to a peak of \textbf{1.43}. RLG thus provides an inference-time 'slider' for alignment strength, a capability static fine-tuning lacks.
\begin{table}[h!]
\centering
\begin{tabular}{l c c}
\toprule
\textbf{Task} & \textbf{$w_{\text{RL}}$} & \textbf{Compression Ratio} \\
\midrule
\multirow{4}{*}{\shortstack{Low Compressibility}} 
& 0.4 & 1.14 \\
& 0.6 & 1.22 \\
& 1.0 & 1.35 \\
& 1.6 & \textbf{1.43}\\
& 3.0 & 1.37 \\
\midrule
\multirow{4}{*}{\shortstack{High Compressibility }} 
& 0.4 & 0.75 \\
& 0.6 & 0.64 \\
& 1.0 & 0.45 \\
& 1.6 & 0.18 \\
& 2.2 & \textbf{0.17} \\
\bottomrule
\end{tabular}
\caption{RLG provides dynamic control over image compressibility. RLG allows for both interpolation and extrapolation beyond the original RL-tuned model's capability ($w_{\text{RL}}=1.0$).}
\label{tab:compressibility_tradeoff}
\end{table}

\paragraph{Balancing Competing Objectives: Text Accuracy vs. Aesthetics.}
Often, maximizing one alignment objective comes at the cost of another. 
Table~\ref{tab:ocr_results} illustrates this conflict. The standard RL-finetuned model ($w_{\text{RL}}=1.0$) achieves a strong OCR accuracy of 88.6\%, but the Aesthetic Score is fixed at 5.20. A user cannot alter this trade-off. RLG transforms this static outcome into a flexible control. For instance, a user prioritizing aesthetics over maximum text accuracy can set $w_{\text{RL}}<1.0$. Conversely, another user can push for peak accuracy at the cost of aesthetics by setting $w_{\text{RL}}=2.8$. This ability to let users choose their preferred sweet spot on the trade-off curve at inference time is a key advantage of RLG that static fine-tuning lacks.

\section{Conclusion and Discussion}
In this paper, we introduced Reinforcement Learning Guidance (RLG), a novel, training-free method for dynamically controlling the alignment of generative models at inference time. RLG allows users to extrapolate beyond and interpolate its baked-in preferences, effectively reducing the KL-regularization penalty to pursue higher rewards. Our extensive experiments demonstrated that RLG consistently enhances performance across a wide array of tasks. RLG offers a simple yet powerful tool for unlocking the full potential of aligned models, providing a flexible control layer over learned preferences.

While RLG has demonstrated broad empirical success, we acknowledge several limitations that suggest avenues for future research.

First, as RLG builds upon the mechanics of CFG, it inherits certain fundamental limitations. Specifically, sampling with CFG scores does not guarantee approximation to the corresponding marginal distribution~\citep{bradley2024classifier, skreta2025feynman}. Thus, while CFG provides a powerful steering mechanism, it exhibits inherent flaws in the subsequent analytical derivations.
Second, our theoretical justification, which links the RLG scale $w$ to the KL-coefficient $\beta$, assumes that the RL-finetuned model has converged to the optimal policy. This is an idealized assumption and only holds when the optimization objective is a standard mixture of expected return and KL divergence. For methods such as GRPO, the optimal policy does not adhere to this theoretical form~\citep{vojnovic2025alignment}. Therefore, further investigation into the mechanism of RLG under broader settings is warranted.
Finally, future work could explore adaptive RLG scales that vary across timesteps or combine RLG with other orthogonal control methods to achieve even more nuanced generation.

\bibliography{aaai2026}
\appendix

\section{Model Specification}
The following table~\ref{tab:model_links} lists the base model, the RL-finetuned model, the reward models and their corresponding links.
\begin{table*}[t]
\centering

\begin{tabular}{ll}
\toprule
\textbf{Model/Reward Function} & \textbf{Link} \\
\midrule
SD3.5-M~\cite{esser2024scalingrectifiedflowtransformers}  & \url{https://huggingface.co/stabilityai/stable-diffusion-3.5-medium} \\
SD1.5~\cite{rombach2022high}  & \url{https://huggingface.co/stable-diffusion-v1-5/stable-diffusion-v1-5} \\
SD1.4~\cite{rombach2022high}  & \url{https://huggingface.co/CompVis/stable-diffusion-v1-4} \\
SDXL-base~\cite{podell2023sdxlimprovinglatentdiffusion}  & \url{https://huggingface.co/stabilityai/stable-diffusion-xl-base-1.0} \\
SD3.5M-FlowGRPO-PickScore~\cite{liu2025flow}  & \url{https://huggingface.co/jieliu/SD3.5M-FlowGRPO-PickScore} \\
SD3.5M-FlowGRPO-Text~\cite{liu2025flow}  & \url{https://huggingface.co/jieliu/SD3.5M-FlowGRPO-Text} \\
SD3.5M-FlowGRPO-GenEval~\cite{liu2025flow}  & \url{https://huggingface.co/jieliu/SD3.5M-FlowGRPO-GenEval} \\
dpo-sd1.5-text2image-v1~\cite{lee2023aligning}  & \url{https://huggingface.co/mhdang/dpo-sd1.5-text2image-v1} \\
SPO-SDXL\_4k-p\_10ep~\cite{liang2024step} & \url{https://huggingface.co/SPO-Diffusion-Models/SPO-SDXL_4k-p_10ep} \\
dpo-sdxl-text2image-v1~\cite{lee2023aligning}  & \url{https://huggingface.co/mhdang/dpo-sdxl-text2image-v1} \\
ddpo-compressibility~\cite{black2023training}  & 
\url{https://huggingface.co/kvablack/ddpo-compressibility} \\
ddpo-incompressibility~\cite{black2023training}  & 
\url{https://huggingface.co/kvablack/ddpo-incompressibility} \\
Aesthetic Score~\cite{LAION-AI_Aesthetic-Predictor} & \url{https://github.com/LAION-AI/aesthetic-predictor} \\
ImageReward~\cite{xu2023imagereward} & \url{https://huggingface.co/THUDM/ImageReward} \\
PickScore~\cite{kirstain2023pickapicopendatasetuser} & \url{https://huggingface.co/yuvalkirstain/PickScore_v1} \\
clip-vit-large-patch14~\cite{radford2021learningtransferablevisualmodels}& \url{https://huggingface.co/openai/clip-vit-large-patch14}\\
stable-diffusion-inpainting~\cite{podell2023sdxlimprovinglatentdiffusion}& \url{https://aihub.caict.ac.cn/models/runwayml/stable-diffusion-inpainting}\\
prefpaint~\cite{bui2025prefpaintenhancingimageinpainting} & \url{https://huggingface.co/kd5678/prefpaint-v1.0}\\
prefpaintreward~\cite{bui2025prefpaintenhancingimageinpainting} & \url{https://huggingface.co/kd5678/prefpaintReward}\\
IP-Adapter-Plus~\cite{ye2023ip} & \url{https://huggingface.co/h94/IP-Adapter}\\
PatchDPO~\cite{huang2025patchdpo} & \url{https://huggingface.co/hqhQAQ/PatchDPO}\\
\bottomrule
\end{tabular}
\caption{Models and Their Corresponding Links.}
\label{tab:model_links}
\end{table*}

\section{Theoretical Derivations}
\label{sec:appendix_prove_RL}

This appendix provides the formal derivations discussed in the main paper.

\subsection{Proof of the Optimal Policy for KL-Regularized RL}
We aim to find the policy $\pi^*$ that solves the optimization problem defined in Equation~\ref{eq:rl_objective}:
\begin{equation}
\pi^* = \arg\max_{\pi} \left( \mathbb{E}_{\mathbf{x} \sim \pi(\mathbf{x})} [R(\mathbf{x})] - \beta D_{\text{KL}}(\pi(\mathbf{x}) \| \pi_{\text{ref}}(\mathbf{x})) \right)
\end{equation}
subject to the constraint that $\pi(\mathbf{x})$ is a valid probability distribution, i.e., $\int \pi(\mathbf{x}) d\mathbf{x} = 1$.

First, we expand the objective functional $J(\pi)$:
\begin{align}
J(\pi) &= \int \pi(\mathbf{x}) R(\mathbf{x}) d\mathbf{x} - \beta \int \pi(\mathbf{x}) \log \frac{\pi(\mathbf{x})}{\pi_{\text{ref}}(\mathbf{x})} d\mathbf{x} \nonumber \\
&= \int \left( \pi(\mathbf{x}) R(\mathbf{x}) - \beta \pi(\mathbf{x}) \log \pi(\mathbf{x}) + \beta \pi(\mathbf{x}) \log \pi_{\text{ref}}(\mathbf{x}) \right) d\mathbf{x}
\end{align}
This is a constrained optimization problem that can be solved using the calculus of variations with a Lagrange multiplier, $\lambda$, for the probability distribution constraint. The Lagrangian is:
\begin{equation}
\mathcal{L}(\pi, \lambda) = J(\pi) + \lambda \left( \int \pi(\mathbf{x}) d\mathbf{x} - 1 \right)
\end{equation}
To find the optimal policy $\pi^*$, we take the functional derivative of $\mathcal{L}$ with respect to $\pi(\mathbf{x})$ and set it to zero.
\begin{align}
\frac{\delta \mathcal{L}}{\delta \pi(\mathbf{x})} &= \frac{\partial}{\partial \pi(\mathbf{x})} \left[ \pi R - \beta \pi \log \pi + \beta \pi \log \pi_{\text{ref}} + \lambda \pi \right] = 0 \nonumber \\
&= R(\mathbf{x}) - \beta(\log \pi(\mathbf{x}) + 1) + \beta \log \pi_{\text{ref}}(\mathbf{x}) + \lambda = 0
\end{align}
Now, we solve for $\log \pi(\mathbf{x})$:
\begin{align}
\beta \log \pi(\mathbf{x}) &= R(\mathbf{x}) + \beta \log \pi_{\text{ref}}(\mathbf{x}) + \lambda - \beta \nonumber \\
\log \pi(\mathbf{x}) &= \frac{1}{\beta} R(\mathbf{x}) + \log \pi_{\text{ref}}(\mathbf{x}) + \frac{\lambda - \beta}{\beta}
\end{align}
Exponentiating both sides gives the form of the optimal policy $\pi^*(\mathbf{x})$:
\begin{align}
\pi^*(\mathbf{x}) &= \exp \left( \frac{1}{\beta} R(\mathbf{x}) + \log \pi_{\text{ref}}(\mathbf{x}) + \frac{\lambda - \beta}{\beta} \right) \nonumber \\
&= \pi_{\text{ref}}(\mathbf{x}) \exp\left(\frac{1}{\beta} R(\mathbf{x})\right) \exp\left(\frac{\lambda - \beta}{\beta}\right)
\end{align}
The term $\exp\left(\frac{\lambda - \beta}{\beta}\right)$ is a constant that does not depend on $\mathbf{x}$. This constant serves as the normalization factor to ensure that $\int \pi^*(\mathbf{x}) d\mathbf{x} = 1$. Let us denote this normalization constant as $1/Z(\beta)$.
Therefore, the optimal distribution is:
\begin{equation}
\pi^*(\mathbf{x}) = \frac{1}{Z(\beta)} \pi_{\text{ref}}(\mathbf{x}) \exp\left(\frac{1}{\beta} R(\mathbf{x})\right)
\end{equation}
This is equivalent to the proportional relationship given in Equation~\ref{eq:rl_dist}:
\begin{equation}
\pi^*(\mathbf{x}) \propto \pi_{\text{ref}}(\mathbf{x}) \exp\left(\frac{1}{\beta} R(\mathbf{x})\right)
\end{equation}
This completes the proof.

\subsection{Equivalence of the DPO Objective}
The Direct Preference Optimization (DPO) framework is derived by re-parameterizing the KL-regularized RL objective in terms of preferences, thereby avoiding the need to explicitly train a reward model. The derivation shows that optimizing the DPO loss is equivalent to optimizing the policy towards the same theoretical distribution $\pi^*$ derived above.

The derivation proceeds as follows:
\begin{enumerate}
    \item \textbf{Express Reward in terms of Policies:} We start with the optimal policy solution from the previous section and rearrange it to solve for the reward function $R(\mathbf{x})$:
    \begin{align}
    \pi^*(\mathbf{x}) &= \frac{1}{Z(\beta)} \pi_{\text{ref}}(\mathbf{x}) \exp\left(\frac{1}{\beta} R(\mathbf{x})\right) \nonumber \\
    \implies R(\mathbf{x}) &= \beta \log\left(\frac{\pi^*(\mathbf{x})}{\pi_{\text{ref}}(\mathbf{x})}\right) + \beta \log Z(\beta)
    \end{align}
    The term $\beta \log Z(\beta)$ is a constant with respect to $\mathbf{x}$.

    \item \textbf{Model Human Preferences:} Human preferences are typically collected as pairs $(\mathbf{x}_w, \mathbf{x}_l)$, where $\mathbf{x}_w$ is preferred over $\mathbf{x}_l$. The Bradley-Terry model maps reward scores to preference probabilities:
    \begin{equation}
    p(\mathbf{x}_w \succ \mathbf{x}_l) = \sigma(R(\mathbf{x}_w) - R(\mathbf{x}_l))
    \end{equation}
    where $\sigma(\cdot)$ is the sigmoid function.

    \item \textbf{Combine Reward and Preference Models:} We substitute the policy-based expression for the reward into the Bradley-Terry model. The constant term $\beta \log Z(\beta)$ cancels out perfectly:
    \begin{align}
    & R(\mathbf{x}_w) - R(\mathbf{x}_l) \nonumber \\
    &= \left(\beta \log\frac{\pi^*(\mathbf{x}_w)}{\pi_{\text{ref}}(\mathbf{x}_w)} + C\right) - \left(\beta \log\frac{\pi^*(\mathbf{x}_l)}{\pi_{\text{ref}}(\mathbf{x}_l)} + C\right) \nonumber \\
    &= \beta \left( \log\frac{\pi^*(\mathbf{x}_w)}{\pi_{\text{ref}}(\mathbf{x}_w)} - \log\frac{\pi^*(\mathbf{x}_l)}{\pi_{\text{ref}}(\mathbf{x}_l)} \right)
    \end{align}
    Thus, the ground-truth preference probability can be expressed entirely in terms of the optimal policy $\pi^*$ and the reference policy $\pi_{\text{ref}}$:
    \begin{equation}
    p(\mathbf{x}_w \succ \mathbf{x}_l) = \sigma\left( \beta \left( \log\frac{\pi^*(\mathbf{x}_w)}{\pi_{\text{ref}}(\mathbf{x}_w)} - \log\frac{\pi^*(\mathbf{x}_l)}{\pi_{\text{ref}}(\mathbf{x}_l)} \right) \right)
    \end{equation}

    \item \textbf{Construct the DPO Loss:} DPO seeks to find a policy $\pi_\theta$ that maximizes the log-likelihood of the observed human preferences. This is equivalent to minimizing the negative log-likelihood loss. By replacing the theoretical optimal policy $\pi^*$ with our parameterized model policy $\pi_\theta$, we arrive at the DPO loss function:
    \begin{equation}
    \mathcal{L}_{\text{DPO}} = - \mathbb{E}_{(\mathbf{x}_w, \mathbf{x}_l) \sim \mathcal{D}} \left( \log p(\mathbf{x}_w \succ \mathbf{x}_l)\right)
    \end{equation}
\end{enumerate}
By minimizing this loss, we are directly training the policy $\pi_{\theta}$ to satisfy the same mathematical relationship that defines the optimal RL policy $\pi^*$. Therefore, the policy obtained by successfully optimizing the DPO objective, $\pi_{DPO}^*$, converges to the same theoretical distribution as the one found by KL-regularized RL, where the reward function $R(\mathbf{x})$ is implicitly defined by the human preference dataset.

\subsection{Derivation of Velocity-Score Relationship}
\label{sec:appendix_v2s_proof}
This section provides a detailed derivation of the relationship between the velocity field $\mathbf{v}(\mathbf{x}, t)$ used in Flow Matching models and the score function $\mathbf{s}(\mathbf{x}, t)$ used in Denoising Diffusion Models, as stated in Equation~\ref{eq:v2s}.

The unifying perspective relies on a common reference path $(X_t)_{t \in [0,1]}$ that interpolates between an initial noise variable $X_1 \sim p_1 = \mathcal{N}(0, \mathbf{I})$ and a data sample $X_0 \sim p_{\text{data}}$. This path is defined by linear interpolation:
\begin{equation} \label{eq:appendix_prob_path}
X_t = \beta_t X_1 + \alpha_t X_0
\end{equation}
where $\alpha_t$ and $\beta_t$ are scalar functions of time, with $\alpha_0 = \beta_1 = 0$ and $\alpha_1 = \beta_0 = 1$.

In Denoising Diffusion Models, the model learns to predict the score function $\mathbf{s}(\mathbf{X}_t, t) = \nabla_{\mathbf{X}_t} \log p_t(\mathbf{X}_t)$. For the chosen linear interpolation path where $X_1 \sim \mathcal{N}(0, \mathbf{I})$, it's a known property that the score function is related to the conditional expectation of $X_0$ and $X_1$ given $X_t$. Specifically, the optimal denoised estimate of $X_0$, denoted $\hat{X}_0(\mathbf{X}_t, t)$, and the optimal estimate of the noise $X_1$, denoted $\hat{X}_1(\mathbf{X}_t, t)$, can be expressed in terms of $X_t$ and its score:
\begin{align}
\hat{X}_0(\mathbf{X}_t, t) &= \mathbb{E}[X_0 | X_t] = \frac{1}{\alpha_t}(\mathbf{X}_t + \beta_t^2 \mathbf{s}(\mathbf{X}_t, t)) \label{eq:appendix_x0_hat} \\
\hat{X}_1(\mathbf{X}_t, t) &= \mathbb{E}[X_1 | X_t] = -\beta_t \mathbf{s}(\mathbf{X}_t, t) \label{eq:appendix_x1_hat}
\end{align}
These relationships hold under the assumption that the conditional distribution $p(X_t | X_0)$ is a Gaussian $X_t \sim \mathcal{N}(\alpha_t X_0, \beta_t^2 \mathbf{I})$, which is implied by the path definition with $X_1 \sim \mathcal{N}(0, \mathbf{I})$.

For Flow Matching models, the objective is to learn a velocity field $\mathbf{v}(\mathbf{X}_t, t)$ that describes the deterministic trajectory of samples via an ordinary differential equation $d\mathbf{X}_t = \mathbf{v}(\mathbf{X}_t, t) dt$. This velocity field matches the time derivative of the reference flow, $\frac{d}{dt}X_t$.
Differentiating Equation~\ref{eq:appendix_prob_path} with respect to time $t$:
\begin{equation} \label{eq:appendix_dxdt}
\mathbf{v}(\mathbf{X}_t, t) = \frac{d}{dt}X_t = \dot{\beta}_t X_1 + \dot{\alpha}_t X_0
\end{equation}
To express the velocity field in terms of the current state $\mathbf{X}_t$ and the score function $\mathbf{s}(\mathbf{X}_t, t)$, we substitute the expressions for $\hat{X}_0(\mathbf{X}_t, t)$ (Equation~\ref{eq:appendix_x0_hat}) and $\hat{X}_1(\mathbf{X}_t, t)$ (Equation~\ref{eq:appendix_x1_hat}) into Equation~\ref{eq:appendix_dxdt}:
\begin{align}
\mathbf{v}(\mathbf{X}_t, t) &= \dot{\beta}_t (-\beta_t \mathbf{s}(\mathbf{X}_t, t)) + \dot{\alpha}_t \left( \frac{1}{\alpha_t}(\mathbf{X}_t + \beta_t^2 \mathbf{s}(\mathbf{X}_t, t)) \right) \nonumber \\
&= -\dot{\beta}_t \beta_t \mathbf{s}(\mathbf{X}_t, t) + \frac{\dot{\alpha}_t}{\alpha_t} \mathbf{X}_t + \frac{\dot{\alpha}_t}{\alpha_t} \beta_t^2 \mathbf{s}(\mathbf{X}_t, t) \nonumber
\end{align}
Rearranging the terms by grouping $\mathbf{X}_t$ and $\mathbf{s}(\mathbf{X}_t, t)$:
\begin{align}
\mathbf{v}(\mathbf{X}_t, t) &= \left( \frac{\dot{\alpha}_t}{\alpha_t} \right) \mathbf{X}_t + \left( \frac{\dot{\alpha}_t}{\alpha_t} \beta_t^2 - \dot{\beta}_t \beta_t \right) \mathbf{s}(\mathbf{X}_t, t) \nonumber \\
&= \left( \frac{\dot{\alpha}_t}{\alpha_t} \right) \mathbf{X}_t + \beta_t \left( \frac{\dot{\alpha}_t}{\alpha_t} \beta_t - \dot{\beta}_t \right) \mathbf{s}(\mathbf{X}_t, t) \label{eq:appendix_v2s_final}
\end{align}
This derivation confirms Equation~\ref{eq:v2s} from the main paper, establishing the precise mathematical connection between the velocity field learned by Flow Matching and the score function predicted by Denoising Diffusion Models under the common linear interpolation path.

\section{Derivation of the Implicit Time-Dependent Reward}
\label{sec:appendix_derive_implicit_reward}
To formalize this, we first need to establish that for any given generative model policy $\pi_{\theta}$ (represented by its distribution $p_{\theta,t}$), we can define a corresponding reward function for which $\pi_{\theta}$ is the optimal policy. This concept is well-established in inverse reinforcement learning for discrete MDPs, such as those used for aligning LLMs~\cite{sun2025f5r, rafailov2023direct}. We can extend this framework to diffusion models by considering the generation process as a continuous-time MDP~\cite{black2023training, rafailov2024rqlanguagemodel}.

In this diffusion MDP, the state at time $t$ is the noisy sample $\mathbf{x}_t$, and the policy $\pi(\cdot|\mathbf{x}_t)$ determines the transition to the next state $\mathbf{x}_{t-dt}$. Recent theoretical work has shown a bijection between reward functions and optimal Q-functions (and thus optimal policies) in such MDPs. Specifically, following~\cite{rafailov2024rqlanguagemodel}, for a given reference policy $\pi_{\text{ref}}$ and a temperature parameter $\beta$, the optimal policy $\pi^*$ for a reward function $r(s_t, a_t)$ satisfies the relationship:
\begin{equation}
    \beta \log \frac{\pi^*(a_t|s_t)}{\pi_{\text{ref}}(a_t|s_t)} = r(s_t, a_t) + \Phi(s_{t-dt}) - \Phi(s_t)
\end{equation}
where $\Phi$ is a potential function, corresponding to the optimal value function $V^*$. This means that any policy $\pi_{\theta}$ can be viewed as the optimal policy for an implicitly defined reward function, equivalent to the log-policy ratio up to a potential-based shaping term.

By adapting this principle to the continuous state-space of diffusion models, we can define an instantaneous, time-dependent reward function $R_t(\mathbf{x}_t)$ directly in terms of the model's probability density. The policy $\pi_{\theta}(\cdot|\mathbf{x}_t)$ is governed by the underlying score function $\mathbf{s}_{\theta}(\mathbf{x}_t, t)$, which itself is the gradient of the log-density $\log p_{\theta,t}(\mathbf{x}_t)$. We can therefore define the implicit reward by relating the marginal densities of the RL-aligned model ($p_{\theta,t}$) and the reference model ($p_{\text{ref},t}$):

\begin{equation}
R_t(\mathbf{x}_t) \triangleq \beta \log \frac{p_{\theta,t}(\mathbf{x}_t)}{p_{\text{ref},t}(\mathbf{x}_t)}
\label{eq:implicit_reward}
\end{equation}
Here, $p_{\theta,t}$ is the marginal probability distribution of the noisy image $\mathbf{x}_t$ under the RL-aligned model, and $p_{\text{ref},t}$ is the corresponding distribution for the pre-trained reference model. The parameter $\beta$ is the same KL-regularization coefficient from the original RL objective (Equation~\ref{eq:rl_objective}). This equation defines the reward that the RL-aligned model $\pi_\theta$ is implicitly optimizing for at every point $(\mathbf{x}_t, t)$ in the generation process, relative to the reference model.

\section{Experimental Details}
\subsection{Generation Hyper-parameters}
\label{sec:appendix_implementation}

This section provides a detailed overview of the parameters and settings used for the text-to-image generation experiments discussed in the main paper.

To ensure consistency across our evaluations, several parameters were standardized for all models. All images were generated at a resolution of $512 \times 512$ pixels. Text prompts were processed using the models' respective text encoders; For the generative process, we uniformly set the number of sampling steps to 20 for all experiments to maintain a balance between computational cost and output quality.

The Classifier-Free Guidance (CFG) scale, which controls the adherence to the text prompt, was set to the generally recommended value for each model to ensure optimal baseline performance. The specific CFG scales used were:
\begin{itemize}
    \item \textbf{Stable Diffusion 1.4:} A CFG scale of 7.5 was used.
    \item \textbf{Stable Diffusion 1.5:} A CFG scale of 7.5 was used.
    \item \textbf{Stable Diffusion inpainting:} A CFG scale of 7.5 was used.
    \item \textbf{Stable Diffusion XL:} A CFG scale of 5.0 was used.
    \item \textbf{Stable Diffusion 3.5:} A CFG scale of 4.5 was used.
\end{itemize}
These CFG scales were held constant across all experiments for a given model, allowing for a direct assessment of the impact of our RLG guidance scale ($w$).
Other parameters are kept default as diffusers pipeline.

\subsection{Human preference Evaluation Metrics Details}
\label{sec:appendix_metrics}

To quantitatively evaluate the performance of our method, we established automated reward models. These models are designed to assess different aspects of image quality and text-to-image alignment, providing a comprehensive evaluation framework.
\begin{itemize}
    \item \textbf{Aesthetic Score}: This metric provides a general assessment of an image's visual appeal. It utilizes a pre-trained CLIP model (`clip-vit-large-patch14`)~\cite{radford2021learningtransferablevisualmodels} to extract a feature embedding from the input image. This embedding is then processed by a Multi-Layer Perceptron (MLP) head, loaded with weights provided in \cite{liu2025flow}, which regresses the features into a single scalar score, typically on a 1-to-10 scale. The corresponding links are listed in table ~\ref{tab:model_links}. A higher score indicates a higher predicted aesthetic quality, as judged by human raters in the dataset the MLP was trained on.

    \item \textbf{ImageReward}: Developed by \cite{xu2023imagereward}, this is a sophisticated reward model designed to evaluate both the semantic alignment of an image to its text prompt and its overall visual fidelity. Built upon the BLIP-2 architecture~\cite{li2023blip2bootstrappinglanguageimagepretraining}, it was fine-tuned on a large-scale dataset of human preference feedback, enabling it to serve as a robust, general-purpose proxy for human judgment in text-to-image generation tasks.

    \item \textbf{PickScore}: Introduced by~\cite{kirstain2023pickapicopendatasetuser}, this reward model is specifically trained to predict human preferences based on direct pairwise comparisons. It is derived from the extensive Pick-a-Pic dataset, which contains a vast number of human choices between two images generated from the same prompt. We use the v1 version, which leverages a powerful `CLIP-ViT-H-14` model~\cite{radford2021learningtransferablevisualmodels}. Its strong correlation with human preference makes it particularly relevant for our work.

\end{itemize}

\section{GenEval Benchmark Details}
\label{sec:appendix_geneval}

The \textbf{GenEval} benchmark provides an automated, object-focused framework for evaluating the compositional capabilities of text-to-image models~\cite{ghosh2023geneval}. Unlike holistic metrics that measure overall image quality or text alignment, GenEval offers a fine-grained analysis of a model's ability to adhere to specific compositional instructions within a prompt.

The official test set is comprised of 553 prompts. the prompts are designed to probe several distinct aspects of compositional generation:

\begin{itemize}
    \item \textbf{Single Object:} Tests the model's fundamental ability to render a single specified object.
    \item \textbf{Two Objects:} Assesses the capacity to generate two distinct objects in the same image, testing for co-occurrence.
    \item \textbf{Counting:} Evaluates whether the model can generate a precise number of a given object.
    \item \textbf{Colors:} Measures if an object can be rendered with a specific, designated color.
    \item \textbf{Position:} Tests spatial reasoning by requiring two objects to be placed in a specified relative position (e.g., "a cat to the left of a dog").
    \item \textbf{Attribute Binding:} The most complex task, which requires binding specific attributes (like color) to specific objects (e.g., generating "a red cube and a blue sphere" without swapping attributes).
\end{itemize}

The evaluation protocol is fully automated, using a sophisticated object detection model to parse the generated images. For our experiments, we adhered to the official methodology, which employs a \textbf{Mask2Former}~\cite{cheng2022maskedattentionmasktransformeruniversal} model with a Swin-S transformer backbone~\cite{liu2021swin}. This detector identifies objects and verifies their properties and spatial arrangements as dictated by the input prompt.

\section{Image Compressibility Experimental Details}
\label{sec:appendix_compressibility}

This section provides a detailed description of the experimental setup for the image compressibility task.

\paragraph{Models and Task Definition.}
The goal of this experiment was to verify that Reinforcement Learning Guidance (RLG) can effectively control a low-level, non-semantic property of generated images: their compressibility. We used the standard Stable Diffusion v1.4 model as our base reference ($\mathbf{v}_{\text{ref}}$). For the aligned models ($\mathbf{v}_{\theta}$), we utilized two sets of weights from the official DDPO implementation~\cite{black2023training}:
\begin{itemize}
    \item \textbf{Low Compressibility Model:} We use ddpo-compressibility. This model was fine-tuned to generate images that are less compressible, resulting in larger file sizes when saved in JPEG format. This typically corresponds to images with higher texture detail and complexity.
    \item \textbf{High Compressibility Model:} We use ddpo-incompressibility. This model was fine-tuned to prefer images that are more compressible, resulting in smaller JPEG file sizes. This often corresponds to images with smoother regions and less high-frequency detail.
\end{itemize}

\paragraph{Dataset and Prompts.}
Following the DDPO study~\cite{black2023training}, our evaluation prompts were based on animal classes from the ImageNet dataset. We used 45 distinct animal classes. For each class, we generated 4 images, resulting in a total of 180 images per RLG scale setting. The prompt template used was: ``\{class\_name\}''.

The 45 animal classes used are: ant, bat, bear, bee, beetle, bird, butterfly, camel, cat, chicken, cow, deer, dog, dolphin, duck, fish, fly, fox, frog, goat, goose, gorilla, hedgehog, horse, kangaroo, lion, lizard, llama, monkey, mouse, pig, rabbit, raccoon, rat, shark, sheep, snake, spider, squirrel, tiger, turkey, turtle, whale, wolf, and zebra.

\paragraph{Evaluation Metric.}
We evaluated performance using the \textbf{Compression Ratio}. For a given prompt, let $\mathbf{x}_{\text{base}}$ be the image generated by the base SD1.4 model, and let $\mathbf{x}_{\text{RLG}}(w)$ be the image generated using RLG with a guidance scale of $w$. Let $S_{jpeg}(\cdot)$ denote the file size of an image after being compressed and saved in JPEG format. The Compression Ratio is defined as:
\[
\text{Compression Ratio}(w) = \frac{S_{jpeg}(\mathbf{x}_{\text{RLG}}(w))}{S_{jpeg}(\mathbf{x}_{\text{base}})}
\]
The final score reported in Table~\ref{tab:compressibility_tradeoff} is the mean of this ratio, averaged across all 180 generated images for each guidance scale. A ratio of 1.0 indicates no change in compressibility compared to the base model.

\section{Image Inpainting Experimental Details}
\label{sec:appendix_inpainting}

This section provides a comprehensive overview of the experimental setup, models, and evaluation protocol used for the image inpainting task discussed in the main paper. Our methodology closely follows that of the PrefPaint~\cite{bui2025prefpaintenhancingimageinpainting} study to ensure a fair and direct comparison.

\subsection{Models and Task Definition}
The experiment focuses on conditional image generation for image inpainting, the task of filling in masked (missing) regions of an image in a semantically and visually plausible manner.

\begin{itemize}
    \item \textbf{Base Model ($\mathbf{v}_{\text{ref}}$):} We use the standard Stable Diffusion inpainting model as our baseline. This model, accessible on diffusers as \texttt{runwayml/stable-diffusion-inpainting}, is widely used and serves as the un-aligned reference point for our experiments.

    \item \textbf{RL-Finetuned Model ($\mathbf{v}_{\theta}$):} For the human-preference-aligned expert model, we employ \textbf{PrefPaint}~\cite{bui2025prefpaintenhancingimageinpainting}. This model is a direct descendant of the base model, which has been further fine-tuned using reinforcement learning. The training process for PrefPaint leveraged a large-scale dataset of over 51,000 human preference judgments on inpainted images, making it an expert policy specialized in generating completions that align with human aesthetic and contextual expectations.
\end{itemize}

\subsection{Evaluation Dataset and Protocol}
All quantitative results were generated using the dataset provided by the authors of PrefPaint~\cite{bui2025prefpaintenhancingimageinpainting}. The test set was constructed using the following procedure:

\begin{enumerate}
    \item \textbf{Image Sourcing:} A diverse set of high-resolution images was initially sourced from established datasets such as ADE20K and ImageNet. All images were resized to a standard $512 \times 512$ pixel resolution.
    \item \textbf{Mask Generation:} To simulate realistic inpainting and outpainting scenarios, two distinct masking strategies were employed to create the incomplete images:
    \begin{itemize}
        \item \textbf{Warping Holes (for Inpainting):} This method creates complex, non-rectangular masks inside the image. It simulates the disocclusion that occurs from a slight change in camera viewpoint. A depth map is first estimated for the source image, and then a new virtual camera view is generated with small shifts. The newly visible (disoccluded) regions form the mask that the model must fill. This tests the model's ability to reason about 3D geometry and handle irregular shapes.
        \item \textbf{Boundary Masks (for Outpainting):} This strategy tests the model's ability to extend a scene beyond its original borders. Masks are created at the edges of the image using two different cropping techniques:
            \begin{itemize}
                \item \textit{Square Cropping:} A central square region, covering 75\% to 85\% of the image area, is preserved, masking the outer frame.
                \item \textit{Rectangular Cropping:} The full height of the image is preserved, while a central vertical slice, comprising 60\% to 65\% of the original width, is kept, masking the left and right sides.
            \end{itemize}
    \end{itemize}
\end{enumerate}

\subsection{Evaluation Metrics}
The quality of the generated inpainted images was assessed using the following two automated metrics:

\begin{itemize}
    \item \textbf{Preference Reward:} We use the specialized reward model developed and released as part of the PrefPaint study~\cite{bui2025prefpaintenhancingimageinpainting}. This model was trained on their custom dataset of nearly 51,000 human preference annotations. Unlike a general-purpose aesthetics model, it is specifically tailored to judge the quality of image inpainting, considering factors like structural rationality, local texture coherence, and overall aesthetic feeling. The reward scores reported in our table are the normalized values from this model, averaged over the official test set, as done in the original paper.

\end{itemize}

\section{Personalized Image Generation Experimental Details}
\label{sec:appendix_personalize}

This appendix provides a detailed overview of the experimental setup for evaluating Reinforcement Learning Guidance (RLG) on the task of personalized image generation, as presented in the main paper.

\subsection{Task and Model Background}
\paragraph{Task Definition.}
Personalized image generation aims to synthesize novel images of a specific subject provided through one or more reference images. The model is given a reference image containing the subject (e.g., a specific pet dog) and a text prompt (e.g., "a photo of [V] sleeping on a couch," where [V] is a placeholder for the subject). The primary goal is to generate an image that not only matches the prompt's description but also maintains high fidelity to the unique appearance and characteristics of the subject in the reference image.

\paragraph{Model Selection.}
Our experiment is designed to test if RLG can amplify the effects of a fine-grained, RL-based alignment process. We therefore select models based on the work of PatchDPO~\cite{huang2025patchdpo}.
\begin{itemize}
    \item \textbf{Base Model ($\mathbf{v}_\text{ref}$):} We use the publicly available, pre-trained \textbf{IP-Adapter-Plus}~\cite{ye2023ip} model built on SDXL~\cite{podell2023sdxlimprovinglatentdiffusion}. IP-Adapter is a powerful method for subject-driven generation that injects image features into the cross-attention layers of a diffusion model. We use it as our baseline because it represents a strong, general-purpose personalization model before any preference-based fine-tuning.
    
    \item \textbf{RL-aligned Model ($\mathbf{v}_\theta$):} We use the model fine-tuned from IP-Adapter-Plus using the \textbf{PatchDPO} algorithm. PatchDPO is a form of preference optimization that operates at a sub-image or "patch" level. During its training, generated images are compared against the reference image. Patches from the generated image that are consistent with the reference subject receive a positive reward, while inconsistent patches are penalized. This process, analogous to reinforcement learning with fine-grained rewards, tunes the model to be highly specialized in preserving subject fidelity.
\end{itemize}

\subsection{Benchmark and Evaluation Metrics}
\paragraph{Benchmark Dataset.}
All evaluations are conducted on \textbf{DreamBench}~\cite{ruiz2023dreambooth}, the standard benchmark for personalized image generation. DreamBench consists of 30 unique subjects, each with a set of reference images and 80 corresponding text prompts. This benchmark is designed to test a model's ability to generate the subject in various contexts, poses, and interactions.

\paragraph{Evaluation Metrics.}
To quantitatively measure the fidelity of the generated images to the reference subject, we employ two standard, complementary metrics. For each prompt in DreamBench, we generate an image and compare it to the ground-truth reference images of the subject.

\begin{itemize}
    \item \textbf{CLIP-I (Image Similarity):} This metric, introduced by the DreamBooth authors, measures the semantic similarity between the generated and reference images. It works by encoding both images into high-dimensional feature vectors using a pre-trained CLIP ViT-L/14 image encoder. The final score is the average cosine similarity between the embedding of the generated image and the embeddings of the reference images. A higher CLIP-I score indicates that the generated image is semantically and stylistically closer to the reference subject from the perspective of the CLIP model.

    \item \textbf{DINO (Structural Similarity):} This metric uses features extracted from a self-supervised ViT-S/16 DINO~\cite{caron2021emerging} model. DINO is trained without labels and learns to capture rich information about object structure, texture, and shape. The metric is calculated as the average cosine similarity between the DINO features of the generated and reference images. It is particularly effective at measuring the preservation of fine-grained details and the structural integrity of the subject, making it an excellent indicator of subject fidelity.
\end{itemize}

\section{Detailed Human Preference Alignment Results}
\label{sec:appendix_aes_full}

This section provides the complete quantitative results for the human preference alignment experiments, complementing the summary presented in Table~\ref{tab:aesthetic_alignment} of the main paper. For each model, we present two tables: one detailing the absolute mean scores for Aesthetic Score, ImageReward, and PickScore across various RLG guidance scales ($w_{\text{RL}}$), and another showing the pairwise win rates against the base model ($w_{\text{RL}}=0.0$) and the standard RL-finetuned model ($w_{\text{RL}}=1.0$).

\begin{table*}[h!]
\centering

\begin{tabular}{c|ccc}
\toprule
\textbf{$w_{\text{RL}}$} & \textbf{Aesthetic Score ($\uparrow$)} & \textbf{ImageReward ($\uparrow$)} & \textbf{PickScore ($\uparrow$)} \\
\midrule
0.0 & 5.97 & 0.99 & 21.75 \\
1.0  & 6.45 & 1.40 & 23.29 \\
1.2 & 6.48 & 1.41 & 23.36 \\
1.4 & 6.54 & 1.40 & 23.48 \\
1.6 & 6.57 & 1.40 & 23.53 \\
1.8 & 6.60 & 1.41 & 23.56 \\
2.0 & 6.62 & 1.40 & 23.57 \\
2.2 & 6.64 & 1.39 & 23.58 \\
2.4 & 6.66 & 1.39 & 23.58 \\
2.6 & 6.68 & 1.37 & 23.59 \\
2.8 & 6.68 & 1.36 & 23.56 \\
\bottomrule
\end{tabular}
\caption{Mean scores for the \textbf{SD3.5-M} model series on human preference metrics across various RLG scales ($w_{\text{RL}}$). The scale $w_{\text{RL}}=0.0$ corresponds to the original \textbf{SD3.5-M} base model, while $w_{\text{RL}}=1.0$ represents the model after GRPO finetuning, named \textbf{SD3.5M-FlowGRPO-PickScore}.}
\label{tab:appendix_sd35_scores}
\end{table*}

\begin{table*}[h!]
\centering

\begin{tabular}{c|ccc|ccc}
\toprule
\multirow{2}{*}{\textbf{$w_{\text{RL}}$}} & \multicolumn{3}{c|}{\textbf{Win Rate vs. Base ($w_{\text{RL}}=0.0$)}} & \multicolumn{3}{c}{\textbf{Win Rate vs. GRPO ($w_{\text{RL}}=1.0$)}} \\
\cmidrule(lr){2-4} \cmidrule(lr){5-7}
& \textbf{Aesthetic} & \textbf{ImageReward} & \textbf{PickScore} & \textbf{Aesthetic} & \textbf{ImageReward} & \textbf{PickScore} \\
\midrule
1.0 & 88.67 & 82.71 & 97.61 & - & - & - \\
1.2 & 89.60 & 83.59 & 98.14 & 57.96 & 53.42 & 60.25 \\
1.4 & 91.31 & 82.71 & 97.80 & 69.29 & 54.44 & 74.95 \\
1.8 & 92.19 & 80.76 & 97.71 & 75.24 & 54.25 & 76.27 \\
2.2 & 92.72 & 78.91 & 97.07 & 77.39 & 53.56 & 73.68 \\
2.4 & 92.87 & 77.54 & 96.78 & 79.10 & 51.66 & 72.80 \\
\bottomrule
\end{tabular}
\caption{Win rates (\%) for \textbf{SD3.5-M} model series at various RLG scales ($w_{\text{RL}}$) compared against the base ($w_{\text{RL}}=0.0$) and GRPO ($w_{\text{RL}}=1.0$) models.}
\label{tab:appendix_sd35_winrates}
\end{table*}

\begin{table*}[h!]
\centering

\begin{tabular}{c|ccc}
\toprule
\textbf{$w_{\text{RL}}$} & \textbf{Aesthetic Score ($\uparrow$)} & \textbf{ImageReward ($\uparrow$)} & \textbf{PickScore ($\uparrow$)} \\
\midrule
0.0 & 6.10 & 0.72 & 21.66 \\
1.0 & 6.42 & 1.12 & 22.69 \\
1.2 & 6.45 & 1.13 & 22.71 \\
1.4 & 6.48 & 1.14 & 22.71 \\
\bottomrule
\end{tabular}
\caption{Mean scores for \textbf{SDXL-base} model series on human preference metrics at various RLG scales ($w_{\text{RL}}$). The scale $w_{\text{RL}}=0.0$ corresponds to the original \textbf{SDXL-base} base model, while $w_{\text{RL}}=1.0$ represents the model after SPO finetuning, named \textbf{SPO-SDXL\_4k-p\_10ep}.}
\label{tab:appendix_sdxl_scores}
\end{table*}

\begin{table*}[h!]
\centering

\begin{tabular}{c|ccc|ccc}
\toprule
\multirow{2}{*}{\textbf{$w_{\text{RL}}$}} & \multicolumn{3}{c|}{\textbf{Win Rate vs. Base ($w_{\text{RL}}=0.0$)}} & \multicolumn{3}{c}{\textbf{Win Rate vs. SPO ($w_{\text{RL}}=1.0$)}} \\
\cmidrule(lr){2-4} \cmidrule(lr){5-7}
& \textbf{Aesthetic} & \textbf{ImageReward} & \textbf{PickScore} & \textbf{Aesthetic} & \textbf{ImageReward} & \textbf{PickScore} \\
\midrule
1.0 & 82.13 & 81.98 & 92.38 & - & - & - \\
1.2 & 83.06 & 81.98 & 92.19 & 59.81 & 54.10 & 54.64 \\
1.4 & 83.64 & 81.20 & 91.46 & 62.99 & 54.15 & 54.35 \\
\bottomrule
\end{tabular}
\caption{Win rates (\%) for \textbf{SDXL-base} model series at various RLG scales ($w_{\text{RL}}$) compared against the base ($w_{\text{RL}}=0.0$) and  SPO ($w_{\text{RL}}=1.0$) models.}
\label{tab:appendix_sdxl_winrates}
\end{table*}

\begin{table*}[h!]
\centering

\begin{tabular}{c|ccc}
\toprule
\textbf{$w_{\text{RL}}$} & \textbf{Aesthetic Score ($\uparrow$)} & \textbf{ImageReward ($\uparrow$)} & \textbf{PickScore ($\uparrow$)} \\
\midrule
0.0 (Base) & 5.51 & -0.02 & 20.03 \\
1.0 (DPO) & 5.61 & 0.20 & 20.39 \\
1.2 & 5.62 & 0.22 & 20.42 \\
1.4 & 5.62 & 0.25 & 20.46 \\
1.6 & 5.64 & 0.26 & 20.51 \\
1.8 & 5.63 & 0.29 & 20.51 \\
2.0 & 5.64 & 0.31 & 20.54 \\
2.2 & 5.64 & 0.31 & 20.55 \\
2.4 & 5.64 & 0.32 & 20.56 \\
\bottomrule
\end{tabular}
\caption{Mean scores for \textbf{SD1.5} model series on human preference metrics at various RLG scales ($w_{\text{RL}}$). The scale $w_{\text{RL}}=0.0$ corresponds to the original \textbf{SD1.5} base model, while $w_{\text{RL}}=1.0$ represents the model after DPO finetuning, named \textbf{dpo-sd1.5-text2image-v1}.}
\label{tab:appendix_sd15_scores}
\end{table*}

\begin{table*}[h!]
\centering

\begin{tabular}{c|ccc|ccc}
\toprule
\multirow{2}{*}{\textbf{$w_{\text{RL}}$}} & \multicolumn{3}{c|}{\textbf{Win Rate vs. Base ($w_{\text{RL}}=0.0$)}} & \multicolumn{3}{c}{\textbf{Win Rate vs. DPO ($w_{\text{RL}}=1.0$)}} \\
\cmidrule(lr){2-4} \cmidrule(lr){5-7}
& \textbf{Aesthetic} & \textbf{ImageReward} & \textbf{PickScore} & \textbf{Aesthetic} & \textbf{ImageReward} & \textbf{PickScore} \\
\midrule
1.0 & 61.52 & 63.33 & 72.66 & - & - & - \\
1.4 & 64.11 & 63.82 & 75.34 & 53.56 & 53.47 & 57.86 \\
1.8 & 63.92 & 66.36 & 76.27 & 55.71 & 56.69 & 61.43 \\
2.2 & 64.55 & 66.46 & 76.22 & 56.40 & 56.64 & 61.72 \\
2.4 & 64.21 & 66.31 & 76.61 & 56.25 & 57.08 & 61.23 \\
\bottomrule
\end{tabular}
\caption{Win rates (\%) for \textbf{SD1.5} at various RLG scales ($w_{\text{RL}}$) compared against the base ($w_{\text{RL}}=0.0$) and standard DPO ($w_{\text{RL}}=1.0$) models.}
\label{tab:appendix_sd15_winrates}
\end{table*}

\section{Selected Image generated}
\subsection{Aesthetic images generated}
\label{sec:aes_image_generated}

\begin{figure*}[h!]
    \centering
    \includegraphics[width=0.9\textwidth]{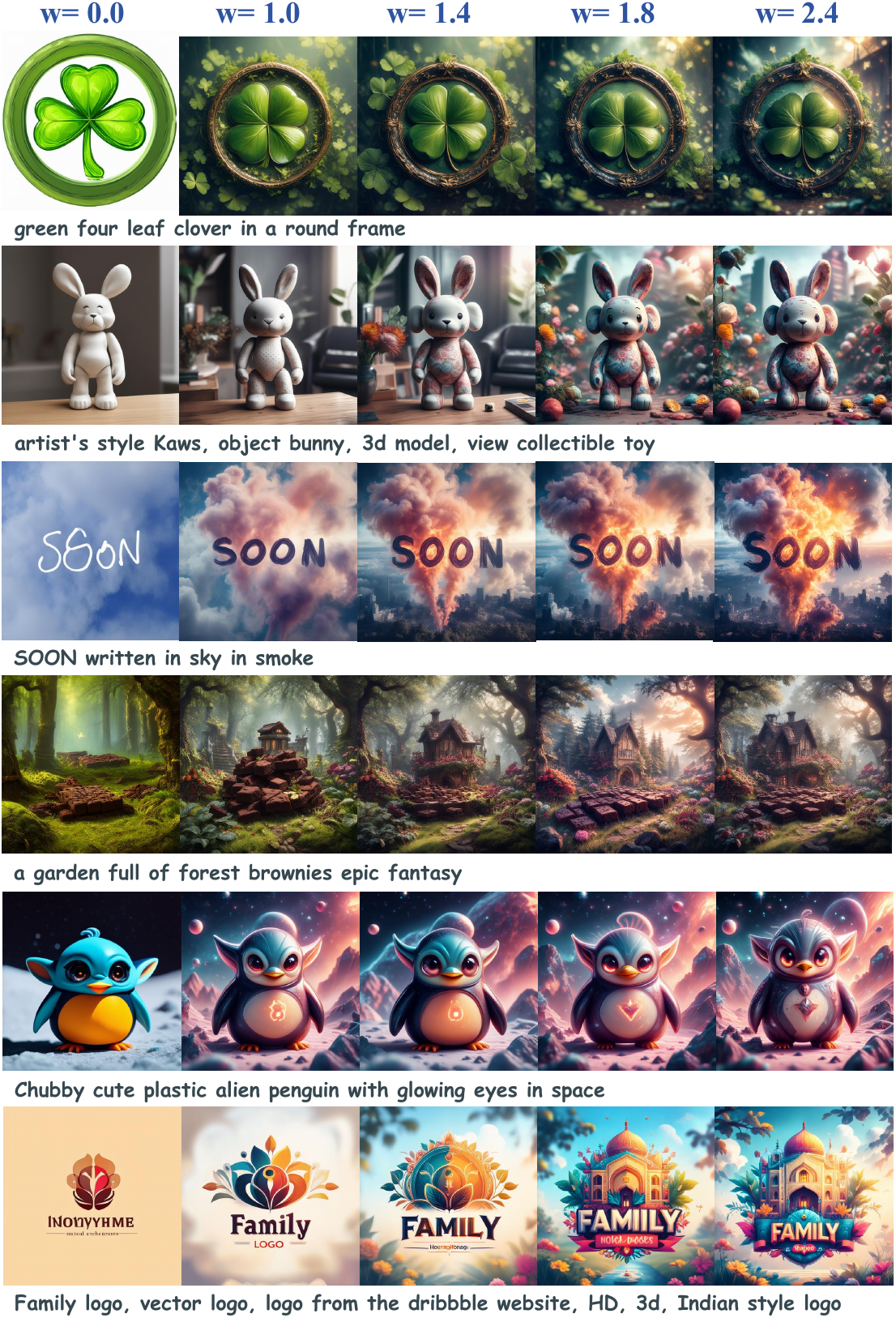}
    \caption{Selected qualitative results for the human preference task. Images are generated from SD3.5 trained with GRPO, with different RLG scales.}
    \label{fig:SD3.5_qualitative}
\end{figure*}

\begin{figure*}[h!]
    \centering
    \includegraphics[width=0.9\textwidth]{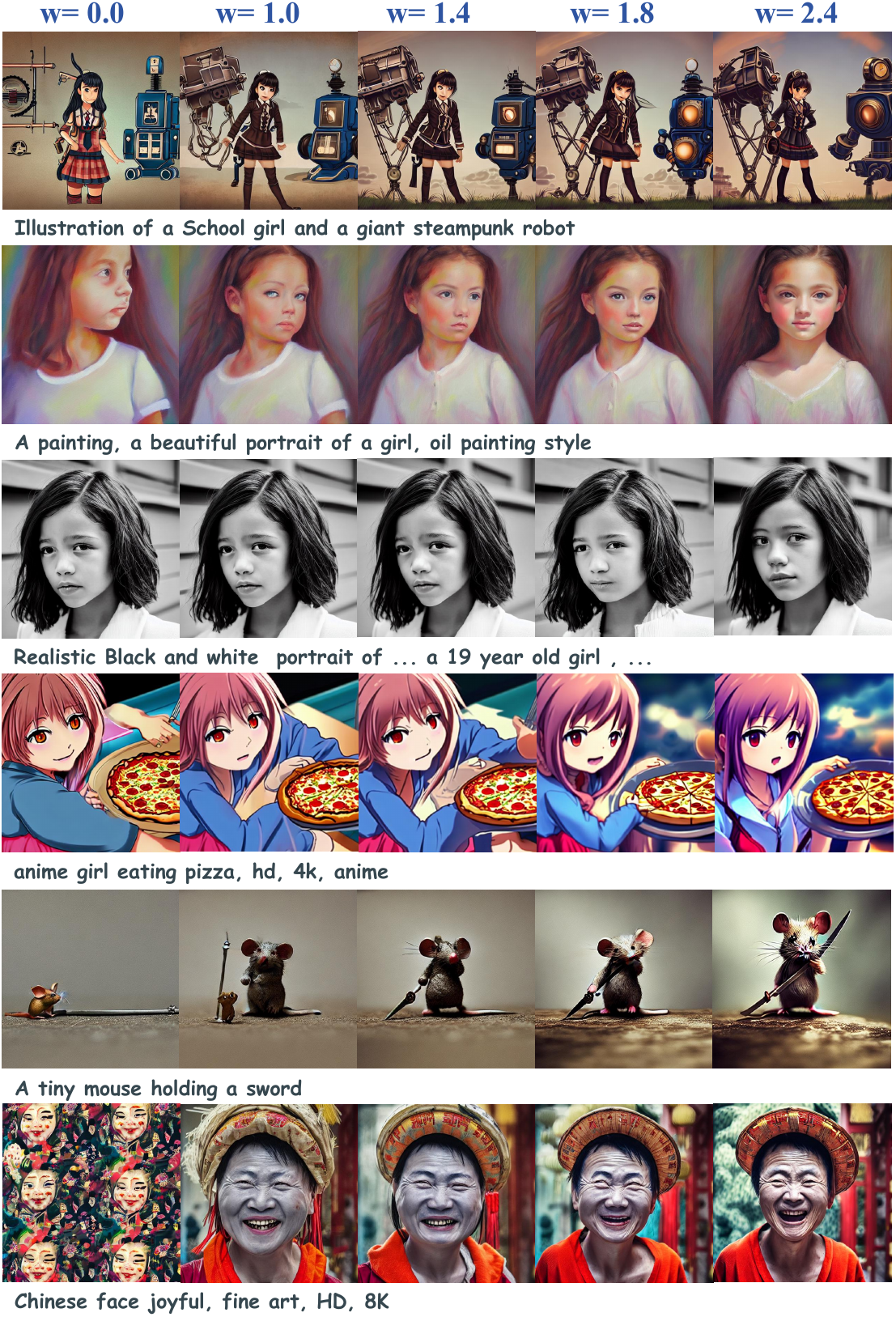}
    \caption{Selected qualitative results for the human preference task. Images are generated from SD1.5 trained with DPO, with different RLG scales.}
    \label{fig:SD1.5_qualitative}
\end{figure*}

\begin{figure*}[h!]
    \centering
    \includegraphics[width=0.9\textwidth]{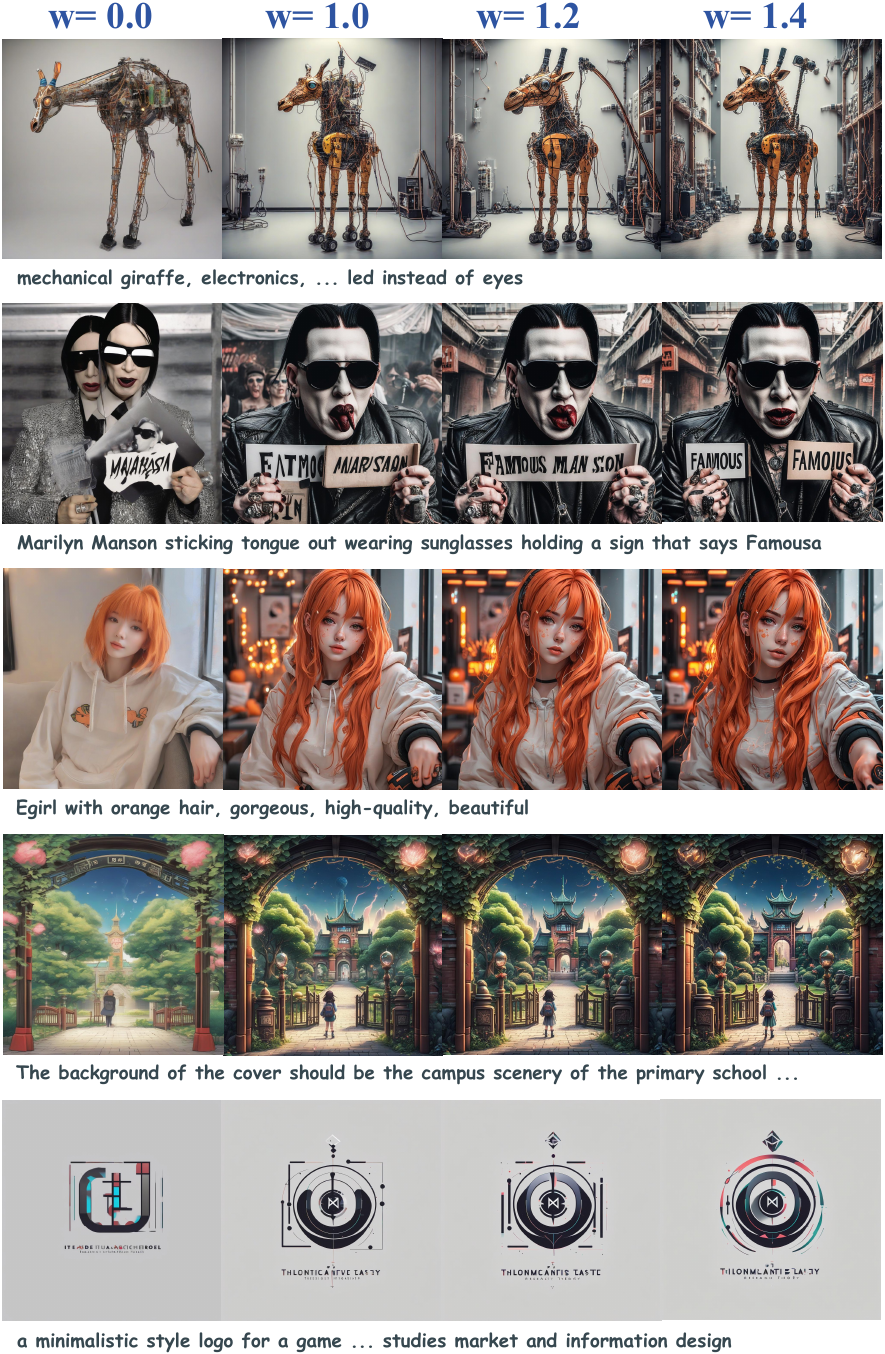}
    \caption{Selected qualitative results for the human preference task. Images are generated from SDXL trained with SPO, with different RLG scales.}
    \label{fig:SDXL_qualitative}
\end{figure*}

\subsection{OCR images generated}
\label{sec:ocr_image_generated}

\begin{figure*}[h!]
    \centering
    \includegraphics[width=\textwidth]{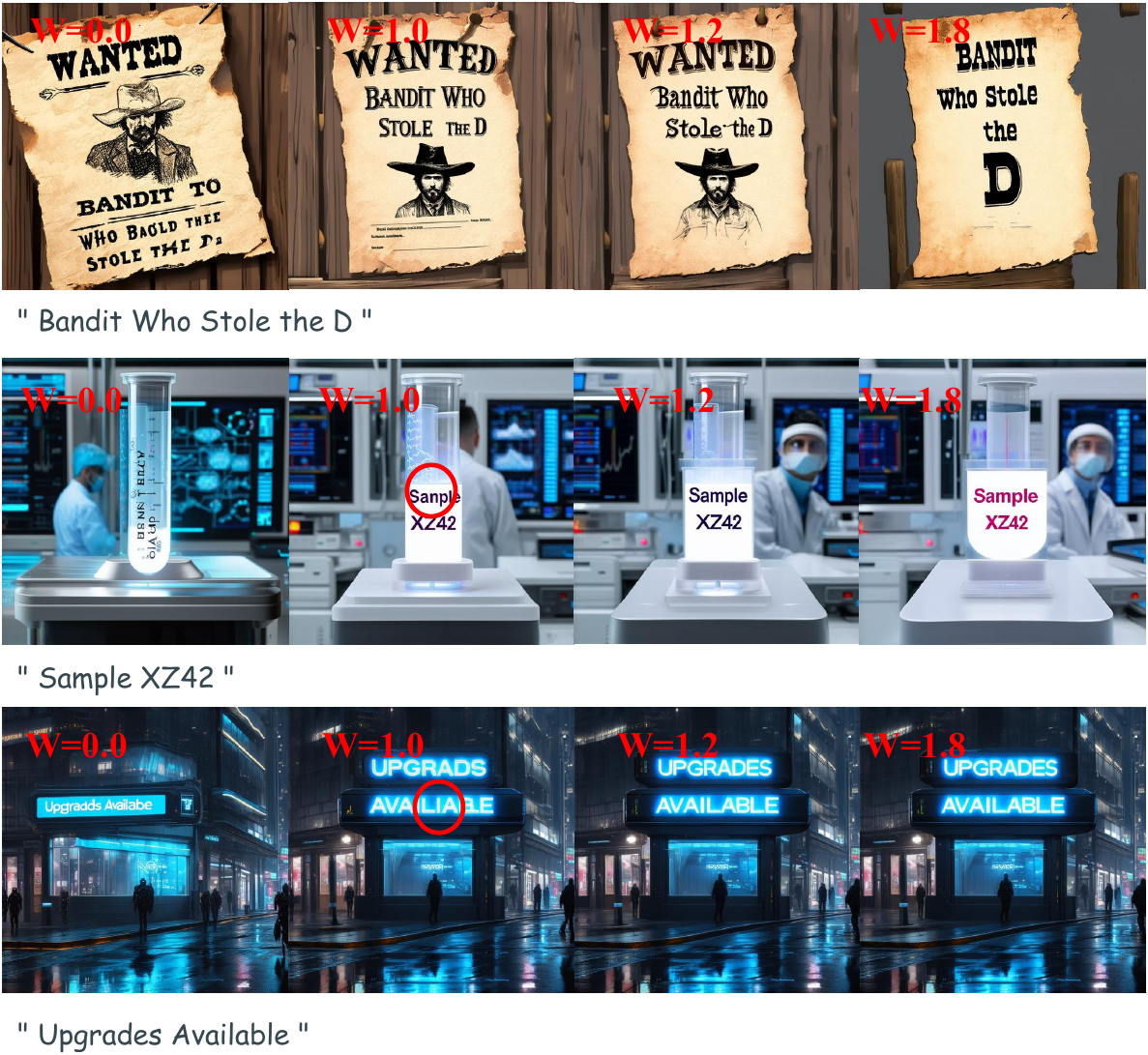}
    \caption{Selected qualitative results for the visual text rendering task. As can be seen, the standard RL-finetuned model ($w=1.0$) still produces some errors in the generated text. By applying RLG with a higher guidance scale ($w>1.0$), the model correctly renders the specified text without any loss in image quality. This illustrates how RLG effectively enhances the model's ability to adhere to precise instructions.}
    \label{fig:ocr_qualitative}
\end{figure*}

\subsection{Compressibility and Incompressibility images generated}
\label{sec:compress_image_generated}

\begin{figure*}[h!]
    \centering
    \includegraphics[width=0.9\textwidth]{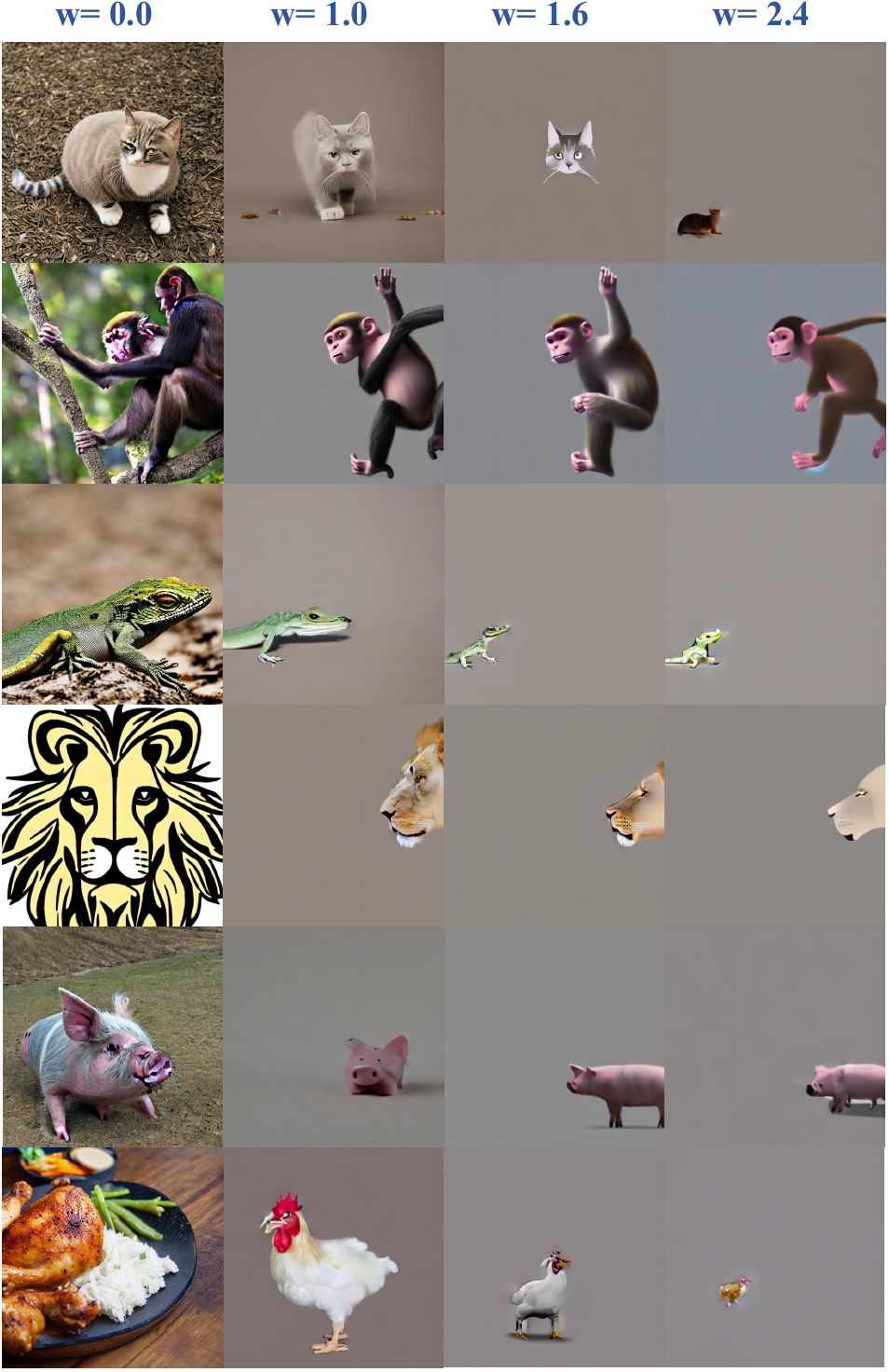}
    \caption{Selected qualitative results for the image compressibility task.}
    \label{fig:compress_qualitative}
\end{figure*}

\label{sec:incompress_image_generated}

\begin{figure*}[h!]
    \centering
    \includegraphics[width=0.9\textwidth]{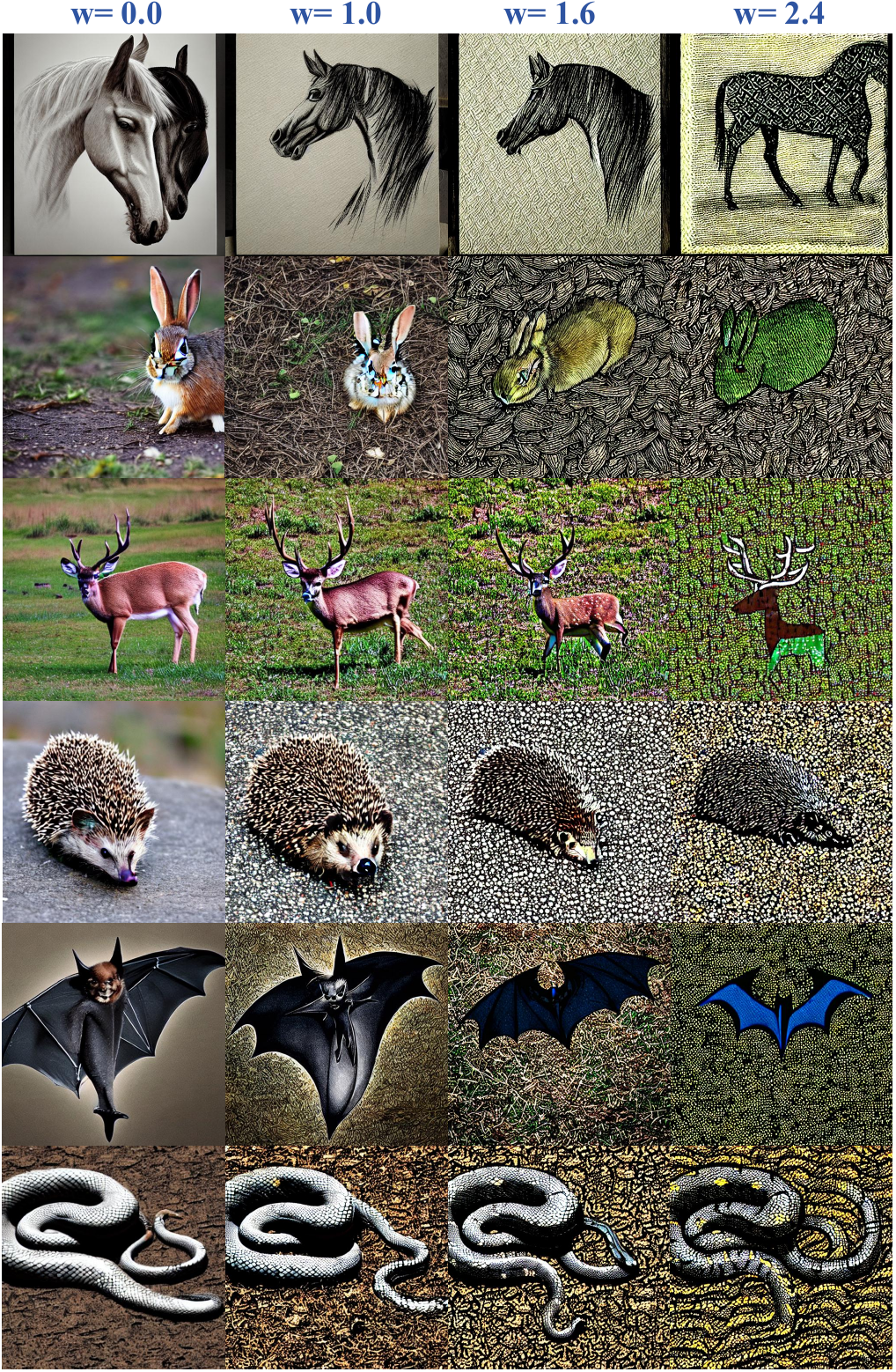}
    \caption{Selected qualitative results for the image compressibility task.}
    \label{fig:incompress_qualitative}
\end{figure*}

\subsection{GenEval images generated}
\label{sec:geneval_image_generated}

\begin{figure*}[h!]
    \centering
    \includegraphics[width=0.9\textwidth]{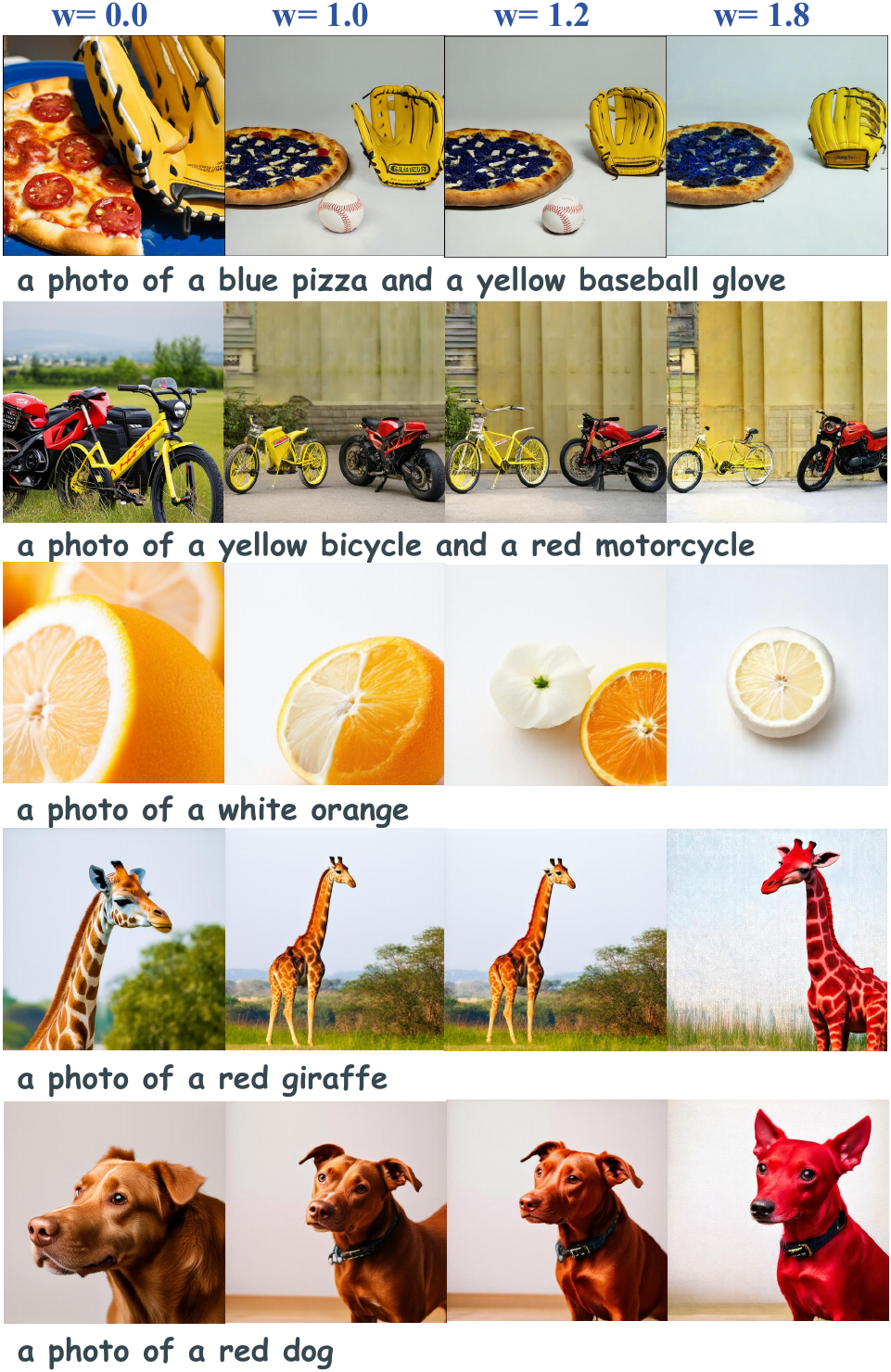}
    \caption{Selected qualitative results for the compositional image generation task.}
    \label{fig:geneval_qualitative}
\end{figure*}

\subsection{Inpainting images generated}
\label{sec:inpaint_image_generated}

\begin{figure*}[h!]
    \centering
    \includegraphics[width=0.9\textwidth]{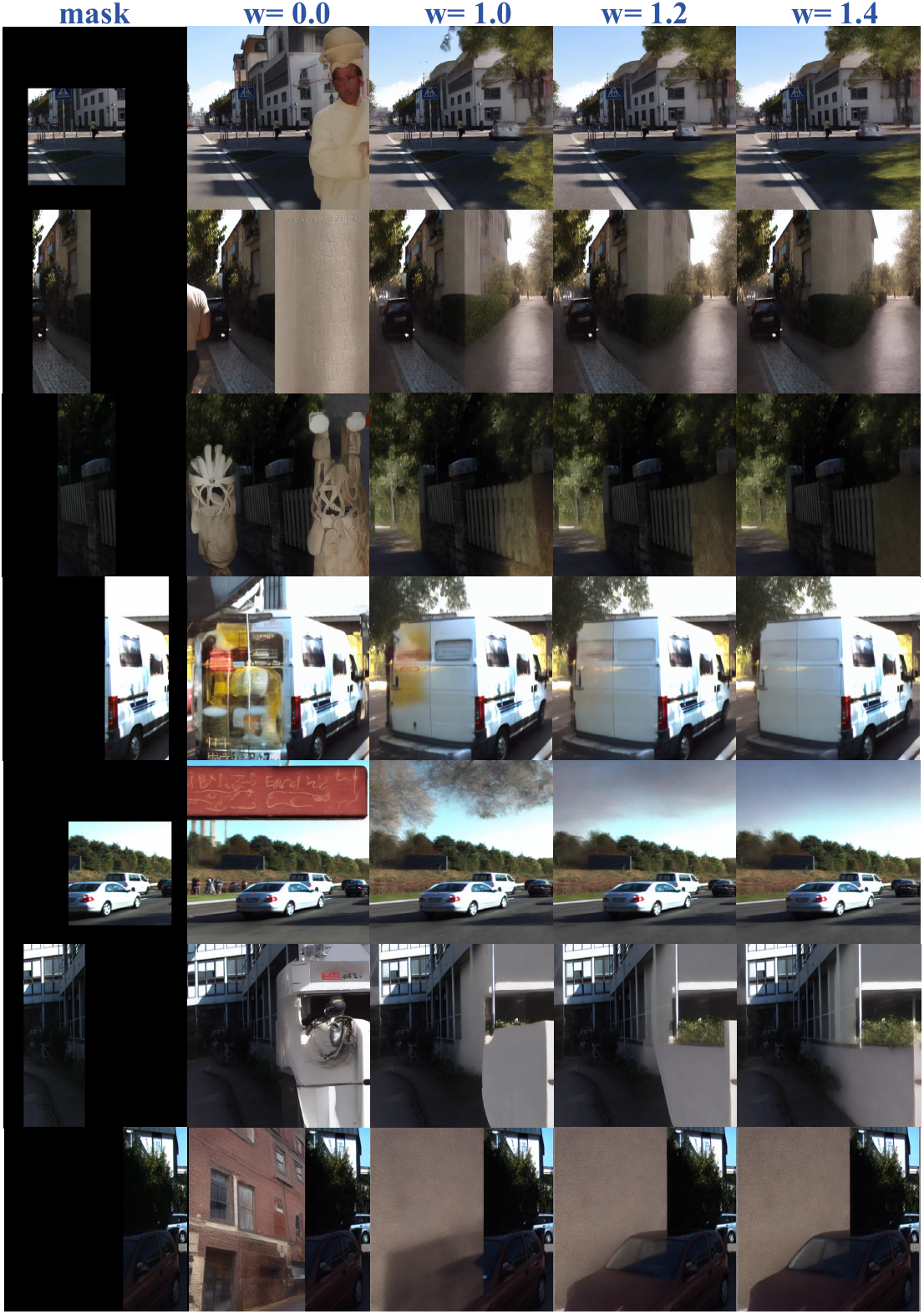}
    \caption{Selected qualitative results for the image inpainting task.}
    \label{fig:inpaint_qualitative}
\end{figure*}

\end{document}